\documentclass[]{sigsicl}
\usepackage{enumitem}
\usepackage[toc,page,header]{appendix}

\usepackage[utf8]{inputenc} 
\usepackage[T1]{fontenc} 
\usepackage{hyperref} 
\usepackage{url}
\usepackage{array} 
\usepackage{booktabs}
\usepackage{amsfonts} 
\usepackage{nicefrac}
\usepackage{microtype}  
\usepackage{xcolor} 
\usepackage{xspace}
\usepackage{bm}
\usepackage{bbm}
\usepackage{bbding}
\usepackage{tabularx}
\usepackage{textcomp}
\usepackage{amssymb}
\usepackage{enumitem}
\usepackage{amsmath}
\usepackage{mathtools}
\usepackage{amsthm}
\usepackage{multirow}
\usepackage{makecell}
\usepackage{color}
\usepackage{colortbl}
\usepackage{adjustbox}
\usepackage{caption}
\usepackage{subcaption} 
\usepackage{graphicx}
\usepackage{wrapfig}
\usepackage{array}
\usepackage{multicol}
\usepackage{lipsum}
\usepackage{float}

\usepackage{marvosym}

\usepackage{fontawesome5}
\usepackage{epigraph}\usepackage[most]{tcolorbox}\definecolor{EmergeQuoteBg}{RGB}{237,246,255}\definecolor{EmergeQuoteBorder}{RGB}{83,148,207}\newtcolorbox{emergequote}{colback=EmergeQuoteBg,colframe=EmergeQuoteBorder,boxrule=0.7pt,arc=3pt,left=8pt,right=8pt,top=7pt,bottom=7pt,before skip=6pt,after skip=10pt,fontupper=\itshape}

\makeatletter
\renewcommand{\paragraph}{%
  \@startsection{paragraph}{4}%
  {\z@}{0.5em}{-0.5em}%
  {\normalfont\normalsize\bfseries}%
}
\makeatother

\definecolor{myyellow}{RGB}{255,192,0}
\definecolor{mygreen}{RGB}{107,170,64}
\definecolor{myred}{RGB}{200,66,25}
\definecolor{mywrite}{RGB}{255,227,132}

\title{%
    \begin{tabular}{@{}c@{\hspace{0.66em}}c@{}}
      \raisebox{-0.22\height}{%
        \includegraphics[height=2.1cm]{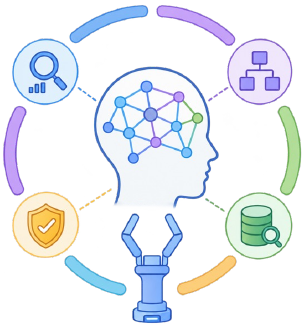}%
      }
      &
      \shortstack[c]{%
      EMERGE-Policy: A Robot Mind \\ Emerges Beyond a Single Policy
      }
    \end{tabular}%
  }

\author{{\bfseries 
Zhirui Fang$^{*,1}$, 
Qingchi Yu$^{*,2,\dagger}$, 
Ziyang Chen$^{*,1}$, 
Longfei Li$^{*,\dagger}$, 
Haoran Ma$^{3,\dagger}$, \\
Keru Zhou$^{1}$, 
Xinrun Xu$^{1}$, 
Samith Va$^{1}$, 
Yuxuan Hu$^{4,\dagger}$, 
Peixuan Song$^{1}$, 
Qiang Du$^{1}$, 
Bin Qian$^{1}$, 
Yongkang Deng$^{5,\dagger}$, 
Xin Li$^{5,\dagger}$, 
Yezhen Wang$^{1}$, 
Zhe Li$^{1}$,
Hao Luo$^{6}$, \\[0.12em]
Shuyan Li$^{1}$,  
Ziwei Wang$^{7}$, 
Weijian Deng$^{1}$, 
Xiu Li$^{1,}$\textsuperscript{\Letter}
}}

\affiliation{{\small
$^{1}$Tsinghua University \quad
$^{2}$Nanjing University of Science and Technology \quad
$^{3}$Xi’an Jiaotong University \\[0.15em]
$^{4}$Xidian University \quad 
$^{5}$Harbin Institute of Technology \quad
$^{6}$Peking University \quad
$^{7}$Nanyang Technological University \\[0.35em]

{\footnotesize
$^{*}$Equal contribution
\hspace{0.9em}
$^{\dagger}$Work done during the internship at Tsinghua University
\hspace{0.9em}
\textsuperscript{\Letter}Corresponding author
}\\[0.55em]

\makebox[\linewidth][c]{%
\resizebox{0.94\linewidth}{!}{%
\begin{tabular}{@{}cccccccc@{}}
\includegraphics[height=0.8cm]{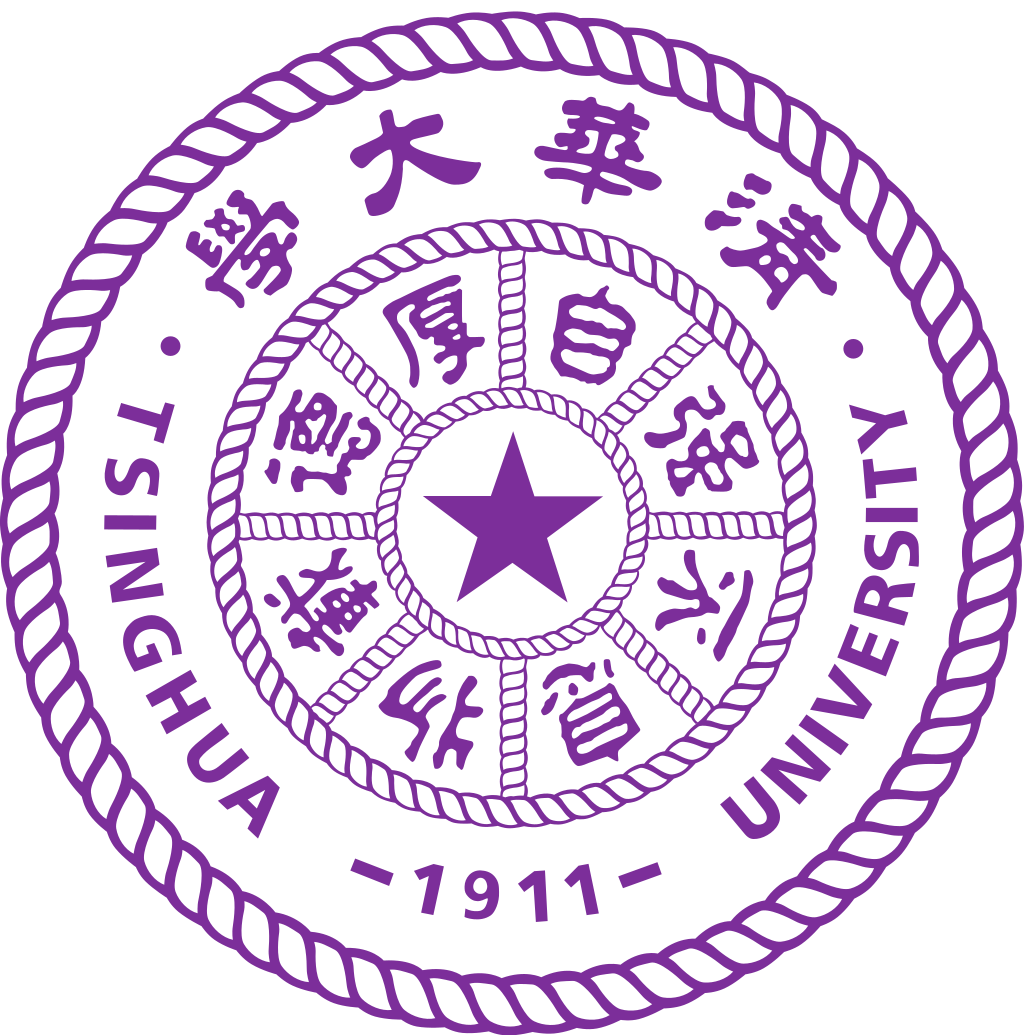} &
\includegraphics[height=0.8cm]{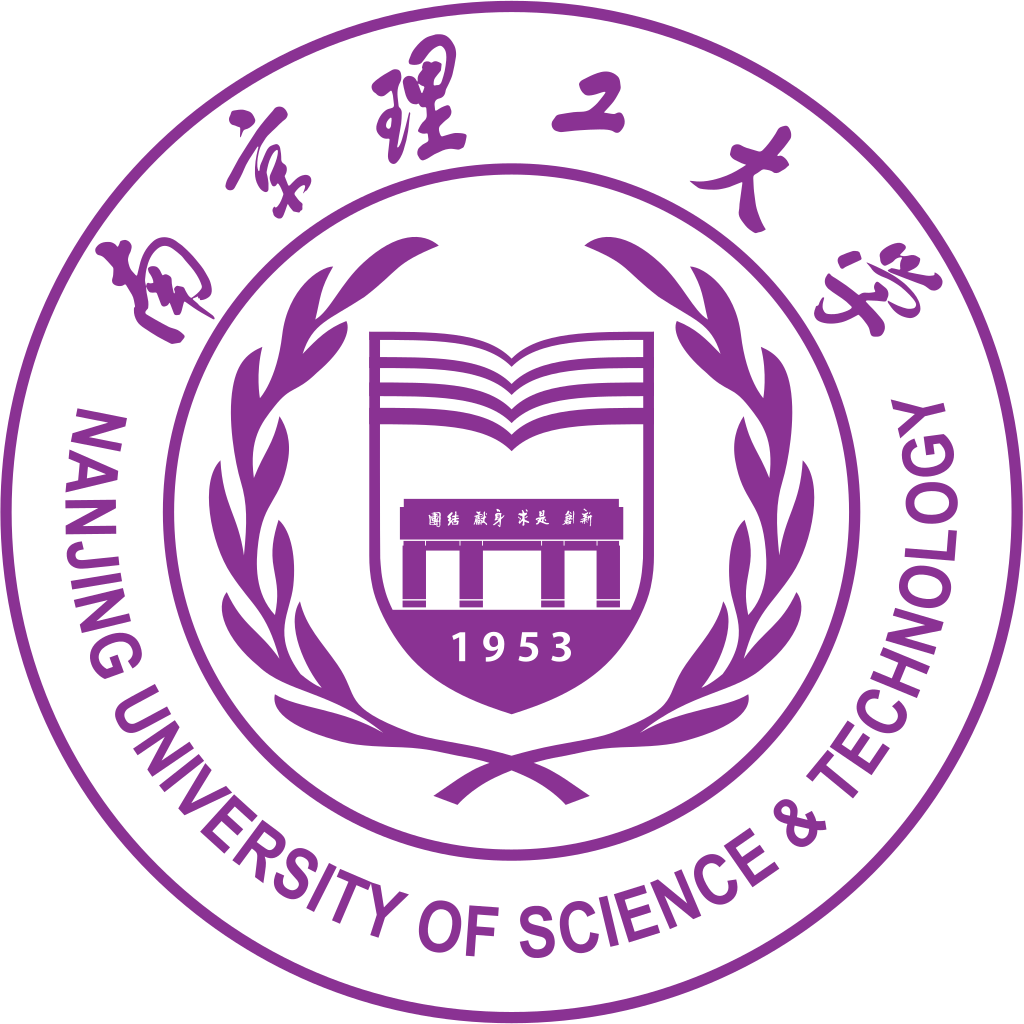} &
\includegraphics[height=0.8cm]{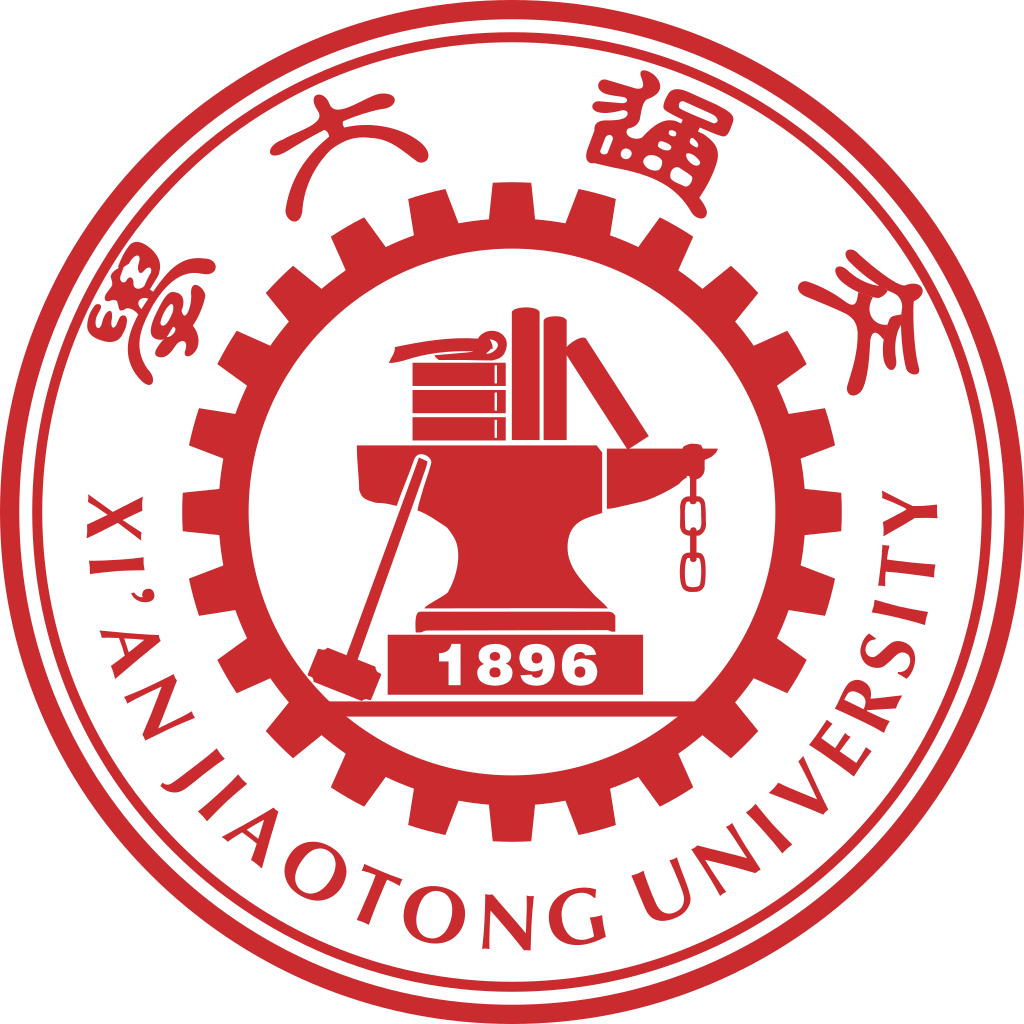} &
\includegraphics[height=0.8cm]{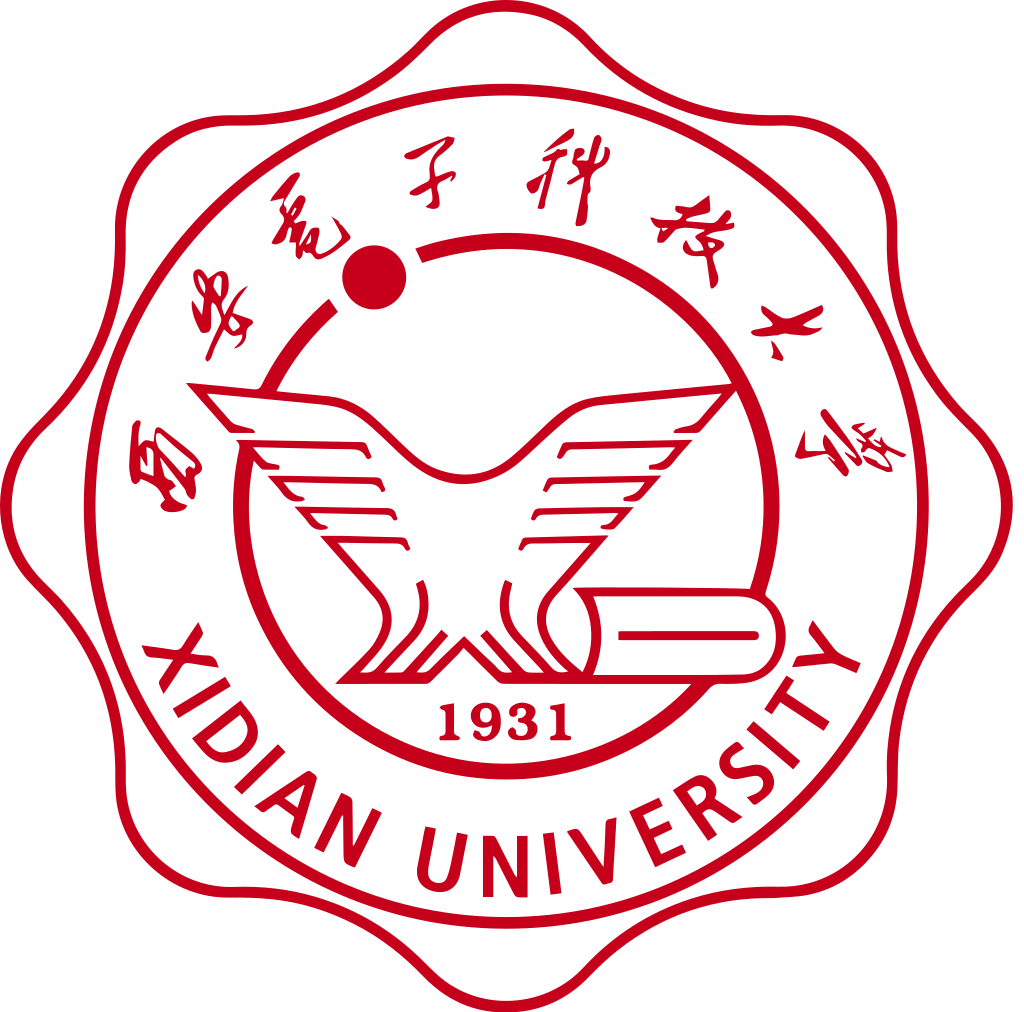} &
\includegraphics[height=0.8cm]{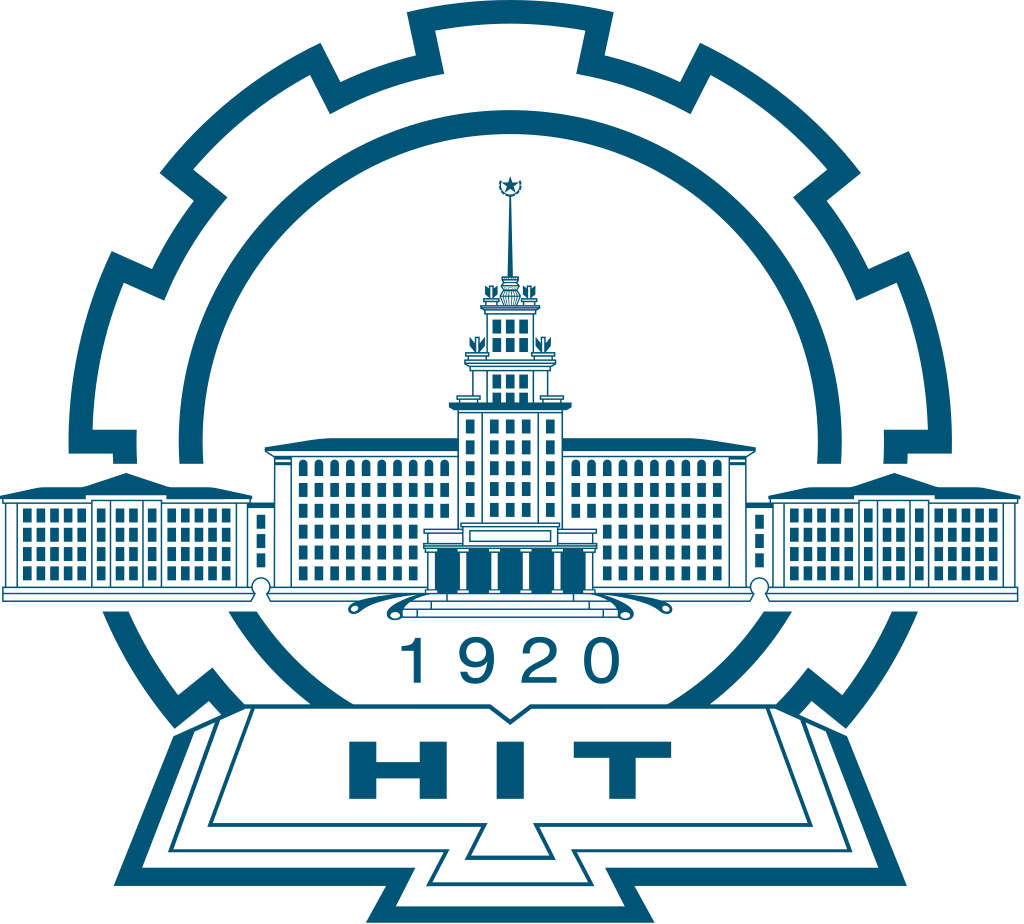} &
\includegraphics[height=0.8cm]{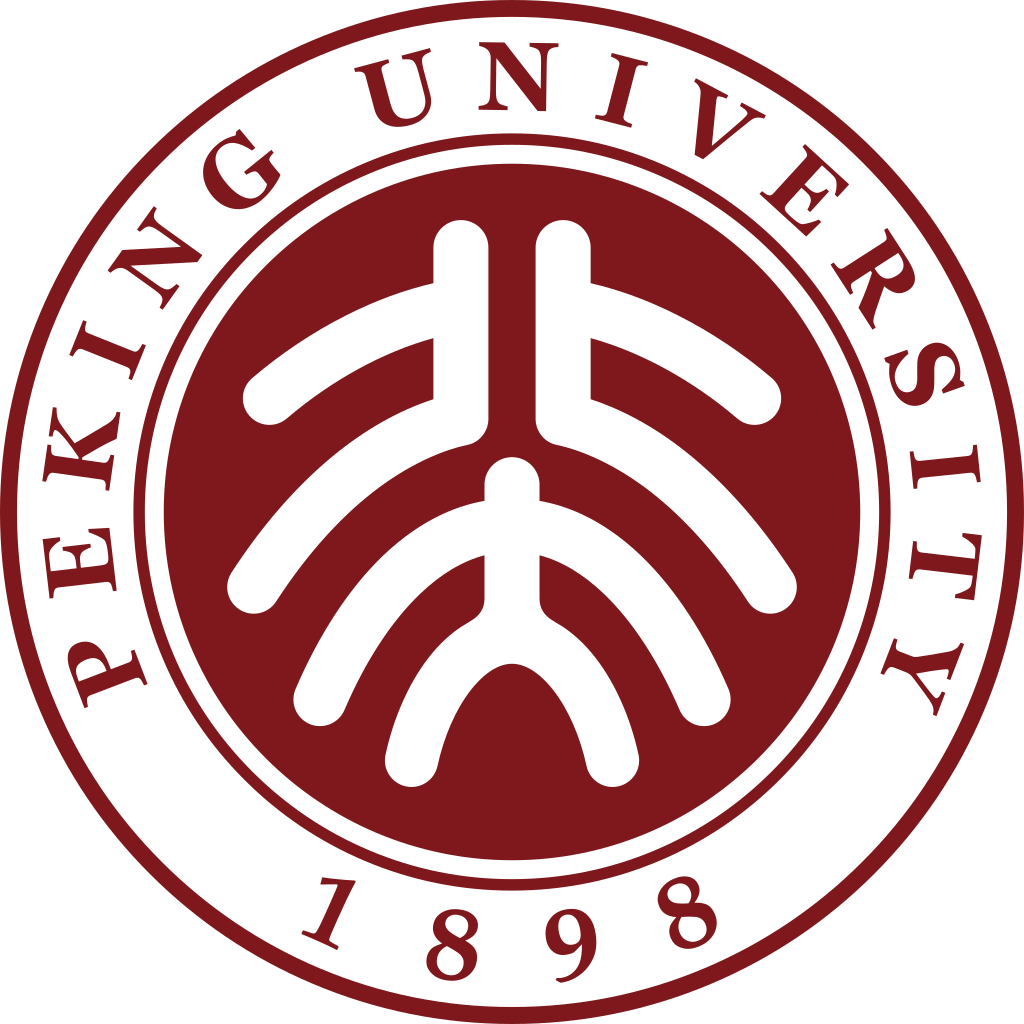} &
\includegraphics[height=0.8cm]{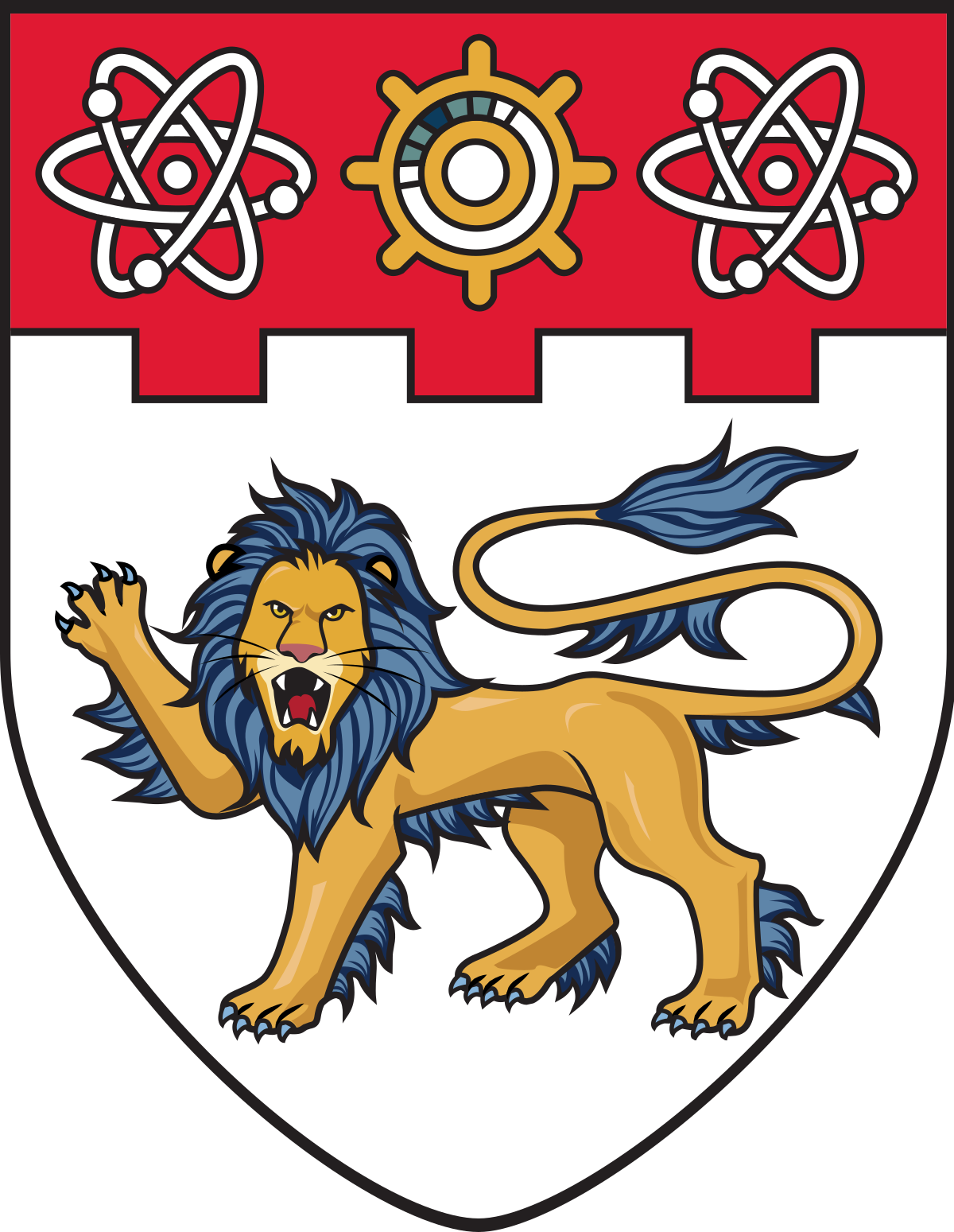}
\end{tabular}%
}}
}}

\webpage{\centering
{\color{blue!70!black}\faLink}\,
\href{https://emerge-policy.github.io/EMERGE-Policy/}{\textbf{Website}}
\hspace{6em}
{\color{black}\faGithub}\,
\href{https://github.com/EMERGE-Policy/EMERGE-Policy}{\textbf{Code}}
}

\firstfig[width=\linewidth][\textwidth]{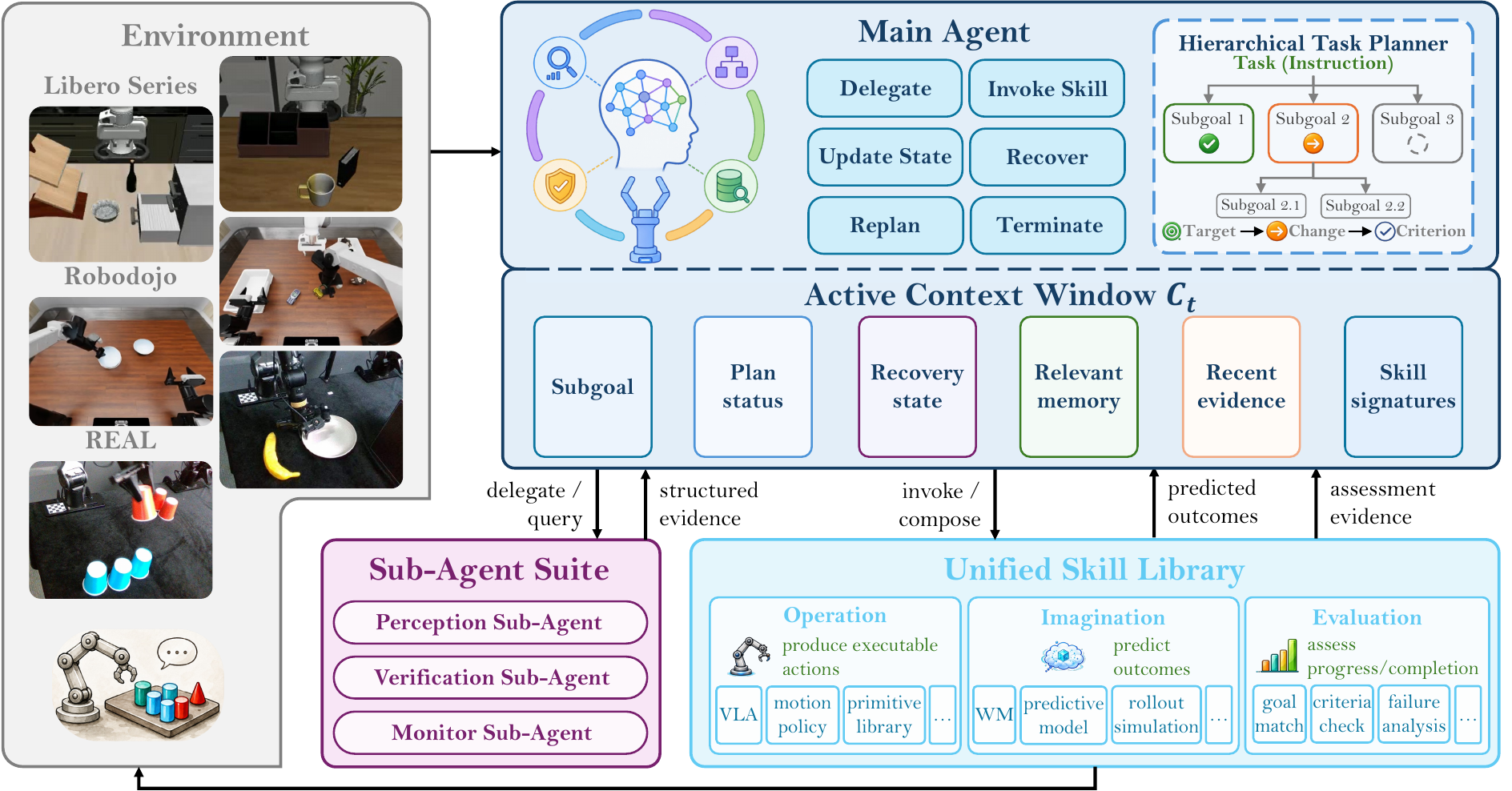}
{{\textbf{System overview of EMERGE-Policy and its coordinated components.}} The Main Agent coordinates Sub Agents with Operational, Imagination, and Evaluation Skills. Structured evidence guides capability selection, verification, and localized recovery across simulation and embodied execution, producing a system-level policy beyond any single backend.}
{fig:001}

\abstract{
{A robot's effective ``mind'' need not reside in a single policy. It can emerge when specialized components perceive, reason, predict, act, verify, and remember within a shared orchestration process. EMERGE-Policy turns this perspective into a graph-structured agentic framework that coordinates both capability invocation and information exchange. A Main Agent retains task-level state within an active context window, while role-specific Sub Agents process perception, execution monitoring, verification, and memory consolidation in isolated contexts and return structured, task-relevant evidence. Role-specific contexts control information load by exposing only decision-relevant evidence to the Main Agent, while the functional Skill interface composes heterogeneous backends as Operational, Imagination, and Evaluation Skills. Criterion-grounded verification, textual failure diagnosis, and Branch Stack recovery provide localized correction, with token-aware external memory preserving task-relevant state. Together, their closed-loop interaction realizes the system-level policy captured by the name EMERGE-Policy. Without additional fine-tuning, we achieved outstanding performance on several public benchmark that have had a wide-reaching impact, and conducted a series of real robot experiments. These system-level results suggest that through the division of different functional sub-tasks among multiple agents and their concurrent collaboration, as well as the technical paradigm where the model is regarded as a skill and called within the framework, EMERGE-Policy can extend the robust robot policies beyond isolated runs.}
}

\definecolor{BlockC}{gray}{0.98}  
\definecolor{BlockA}{RGB}{191,211,230}
\definecolor{BlockB}{RGB}{199,233,192}

\begin{document}

\maketitle

\section{Introduction}
\label{sec:introduction}

\begin{emergequote}
``Thinking outside the brain means skillfully engaging entities external to our heads.''

\par\smallskip\hfill ---Annie Murphy Paul, \textit{The Extended Mind}~\citep{paul2021extended}

\end{emergequote}

  {For embodied agents, the long-term goal pursued by practitioners is \textbf{how to construct a human-like and efficient embodied intelligent system}, this observation has a concrete systems interpretation: intelligence is not confined to a single policy or reasoning model, but is realized through the coordinated interaction of perception, prediction, task reasoning, physical control, external memory, and feedback from the environment. This coordination becomes particularly consequential in robot manipulation, where a robot must repeatedly ground an instruction in the current scene, select and execute an appropriate behavior, verify the resulting physical state, and respond when execution deviates from the plan. Because an error at one stage can invalidate the assumptions of subsequent stages, reliable behavior depends not only on the capability of individual components, but also on how information and control are organized across the complete system. The central challenge is therefore to coordinate specialized capabilities while controlling how much information the high-level planner must process at each decision step.}

  {Existing work provides several complementary components for this process. Vision-Language-Action models have progressed from transferring pretrained vision-language representations to robot control~\citep{kim2024openvla,brohan2023rt}, through diffusion- and flow-based action generation~\citep{li2024cogact,reuss2024multimodal,black2024pi_0,intelligence2025pi_,pertsch2025fast}, to dual-system designs that combine deliberative reasoning with reactive control~\citep{bjorck2025gr00t}. Video-based pretraining introduces broader priors over human activities and physical dynamics~\citep{cheang2024gr,shen2026videovla,hu2024video}, while World-Action Models couple action generation with future-state prediction~\citep{kim2026cosmos,ye2026world,yuan2026fast}. In parallel, Code as Policies~\citep{liang2023code} and RoboCodeX~\citep{mu2024robocodex} represent robot behavior as executable programs over perception and control APIs; CaP-X~\citep{fu2026cap} studies how abstraction, interaction, and grounding affect coding-agent reliability, and ASPIRE~\citep{lu2026aspire} converts validated repairs into reusable skills. More recently, RoboClaw~\citep{li2026roboclaw}, HarnessVLA~\citep{zhang2026harness}, and RoboHarness~\citep{huang2026roboharness} demonstrate how high-level agents can orchestrate learned policies, analytical controllers, perceptual tools, verification, and execution memory. Although these systems differ in scope, their orchestration is commonly centered on capability selection and control transfer. Detailed outputs from perception, execution monitoring, verification, and memory processing can therefore be passed directly to the high-level planner, even when only part of that information is relevant to its next decision. This motivates processing such evidence separately and returning structured, decision-relevant results to the planner.}

  {We introduce \textbf{EMERGE-Policy}, a graph-structured agentic policy framework for robot manipulation. Its name reflects how coherent agentic behavior can emerge from the interaction of specialized components rather than being assigned to a single model. This emergence is structured rather than incidental: EMERGE-Policy makes functional roles, information boundaries, capability invocation, verification, and recovery explicit, so that the coordinated system acts as a policy beyond any one component. A \emph{Main Agent} maintains compact global task state, decomposes instructions into criterion-grounded subgoals, and coordinates the execution process. Specialized \emph{Sub Agents} handle perception, execution monitoring, verification, and memory consolidation in isolated contexts, returning structured evidence rather than complete interaction histories. This organization limits interference from dense low-level evidence while retaining the information required for task-level decisions. By making evidence boundaries explicit, it also clarifies which observation or result informs each orchestration decision.}

  {EMERGE-Policy connects this information-flow organization to action through a shared subgoal specification. Each subgoal defines a target entity, an intended state change, and an explicit completion criterion, providing a common contract for three complementary Skill roles. \emph{Operational Skills} execute actions toward the intended change, \emph{Imagination Skills} predict candidate outcomes, and \emph{Evaluation Skills} assess feasibility, progress, and criterion satisfaction. Because these roles are defined by function rather than implementation, different policy and world-model backends can be composed within the same closed loop. After execution, a satisfied criterion advances the plan; otherwise, the system generates a textual diagnosis and pushes a localized recovery objective onto the \emph{Branch Stack}. Once recovery succeeds, execution returns to the interrupted plan, while global replanning is reserved for failures that local correction cannot resolve. Token-aware external memory preserves task state and relevant evidence throughout this process. Together, these mechanisms control information load, compose heterogeneous capabilities, and localize correction; their closed-loop interaction produces the system-level policy captured by EMERGE-Policy.}

  {We evaluate EMERGE-Policy on standard and perturbed LIBERO benchmarks and study its deployment in a multistep real-robot manipulation task. The experiments examine the complete orchestration system with different VLA and WAM execution backends, particularly in settings involving semantic grounding, outcome verification, and recovery. 
}
  {Our contributions are threefold:}
\begin{itemize}[leftmargin=*]
    \item   {We introduce EMERGE-Policy, a graph-structured agentic framework in which robot behavior emerges from the coordination of a Main Agent, role-specific Sub Agents, Skills, and external memory rather than from a single policy. By adopting this approach, we divided the task into several parts. Different agents can independently complete the sub-tasks or sub-functions, thereby solving the overall problem.}
    
    \item   {We have expanded the definition of skills. EMERGE-Policy regards the currently widely-discussed models such as VLA, World model, and Verifier model as the skills of the agent. These skills are invoked by the agent at different stages of completing the task. Furthermore, we define explicit information and control interfaces: Sub Agents return only structured, task-relevant evidence, while functional Skills, completion criteria, Branch Stack recovery, and episodic memory connect planning, action, and feedback.}

    \item   {We conducted thorough and extensive experiments on a series of widely adopted benchmarks. The experiment shows that our framework has achieved a significant improvement compared to the most relevant models ($\pi_{0.5}$ and Cosmos Policy). It has increased by 2.0\% and 0.7\% on LIBERO, by 2.6\% and 11.7\% on LIBERO-plus, and significantly enhanced the performance of memory on Robodojo, and increased the success rate by 3.59\% compared to the baseline, and we deploy it on a physical robot for a series of long-horizon tasks.}
\end{itemize}

\section{EMERGE-Policy: Graph-Structured Agentic Policy Orchestration}
\label{sec:Methods}

\begin{figure}[H]
    \centering
    \includegraphics[width=1.0\textwidth]{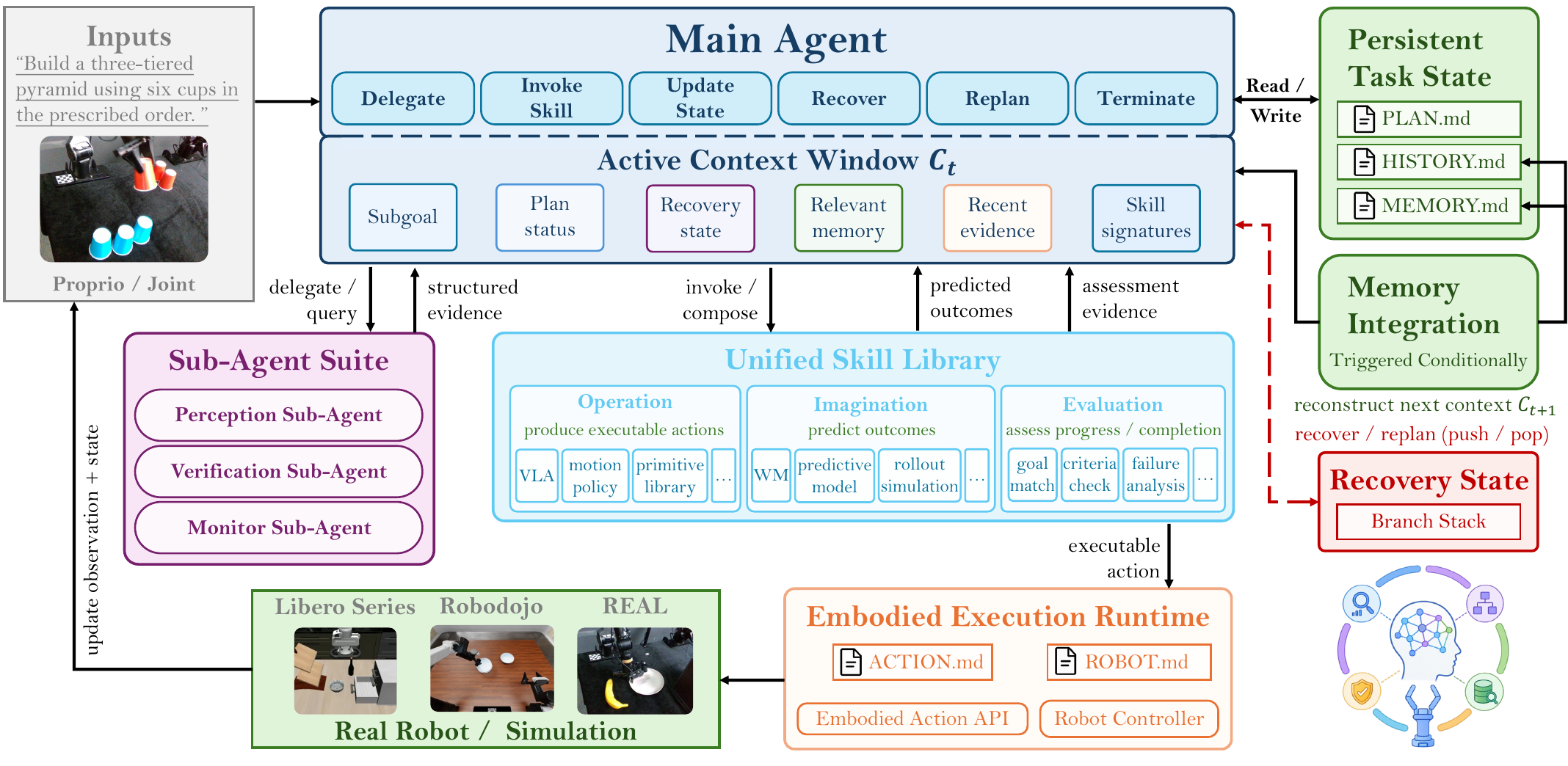}
    \caption{{\textbf{EMERGE-Policy workflow from instruction to verified execution.} The Main Agent maintains active task context and coordinates role-specific Sub Agents with Operational, Imagination, and Evaluation Skills. Structured evidence and Skill outcomes update task state after each interaction: verified successes advance the plan, while failures enter localized Branch Stack recovery before global replanning. Episode memory preserves relevant task state throughout execution.}}
    \label{fig:framework_method}
\end{figure}

\paragraph{Overview.}
 {As shown in \Cref{fig:framework_method}, given a language instruction $\mathcal{T}$ and the current observation $\mathcal{O}_t$, EMERGE-Policy organizes robot manipulation as closed-loop coordination among a Main Agent, role-specific Sub Agents, and a unified Skill library. The Main Agent maintains an active task context $\mathcal{C}_t$, decomposes the instruction into subgoals with explicit completion criteria, and determines which Sub Agent or Skill to invoke next.}

 {Sub Agents process task-specific evidence for perception, execution monitoring, verification, and memory consolidation in isolated contexts, returning concise structured results to the Main Agent. The Skill library organizes heterogeneous capabilities by their roles in the decision loop: Operational Skills execute robot actions, Imagination Skills predict candidate outcomes, and Evaluation Skills assess feasibility, progress, or completion. Because Skills share this functional interface, heterogeneous backends can participate in the same orchestration protocol without assuming identical internal architectures.}
 {Following each physical execution, the observed outcome is evaluated against the active subgoal criterion. Successful verification advances the plan, whereas failure introduces a localized recovery subgoal into the Branch Stack. If local recovery remains unsuccessful, the Main Agent revises the high-level plan. External memory preserves task state and relevant evidence throughout this closed-loop process. Within this graph, role-specific contexts regulate information flow, the Skill interface composes heterogeneous backends, and criterion-based verification with the Branch Stack supplies localized correction. The system-level policy emerges from their closed-loop interaction, which determines how each module's action or evidence shapes the next decision.}


\subsection{Main Agent: Planning, Recovery, and Memory
}
\label{Main Agent}

\begin{figure}
    \centering
    \includegraphics[width=1\linewidth]{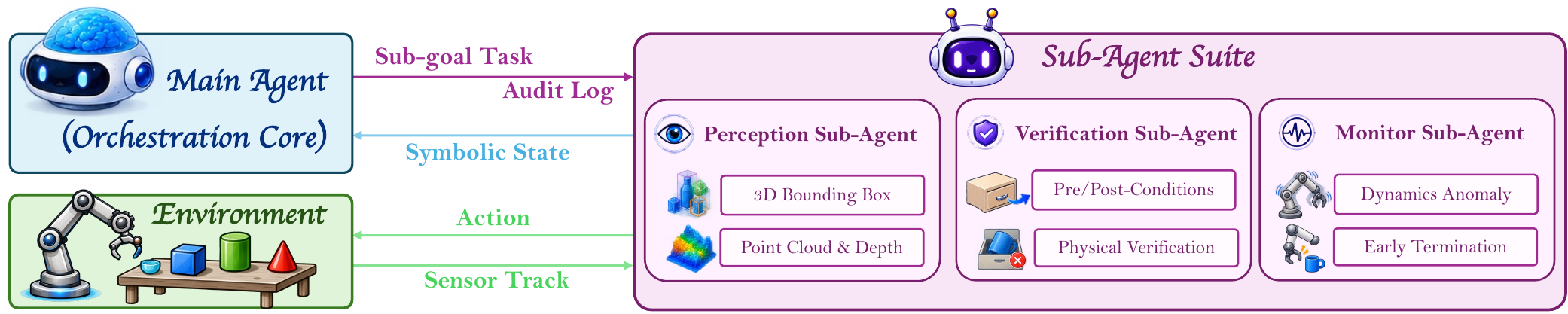}
    \caption{{\textbf{Main Agent orchestration and feedback flow.} The Main Agent selects task-level decisions from an active context containing the instruction, plan status, recovery state, relevant memory, recent evidence, and compact Skill signatures. It delegates evidence processing to Sub Agents, invokes Skills, and updates or revises the plan. Structured evidence and verification outcomes return to the active context, advancing the task or triggering localized recovery.}}
    \label{fig:mian_and_sub}
\end{figure}

 {The Main Agent coordinates task-level planning, capability invocation, outcome verification, and failure recovery. Given the instruction $\mathcal{T}$ and evidence derived from the current observation $\mathcal{O}_t$, it maintains an active task context $\mathcal{C}_t$. The Main Agent implements an LLM-based orchestration policy $\pi_{\mathrm{M}}$ over a decision space $\mathcal{D}$. At interaction step $t$, the orchestration decision is given by}
\begin{equation}
    d_t \sim
    \pi_{\mathrm{M}}
    \left(
        \cdot \mid \mathcal{C}_t
    \right),
    \qquad d_t \in \mathcal{D}.
\end{equation}
 {Here, $\mathcal{D}$ comprises decisions to delegate evidence processing to a Sub Agent, invoke a Skill, update the task state, initiate recovery, revise the plan, or terminate execution. See the \Cref{fig:mian_and_sub} for details.}

\subsubsection{Hierarchical Context Engineering}

 {The active context $\mathcal{C}_t$ contains the task instruction, current subgoal, plan status, top Branch Stack entry, task-relevant memory, recent structured results, and compact signatures of the available Skills. Raw multimodal observations and complete execution traces are kept outside this context unless required for the current decision.}
 {EMERGE-Policy controls context growth in three ways. First, stable constraints and embodiment configurations are separated from dynamic task state and inserted only when relevant. Second, the Main Agent initially receives only compact Skill signatures; complete parameter schemas are retrieved when a Skill is selected for invocation. Third, evidence-intensive perception, monitoring, verification, and memory operations are delegated to Sub Agents operating in isolated contexts. Each Sub Agent receives a task-specific request and returns a structured result rather than its complete processing history.}




\subsubsection{Hierarchical Task Planner and Branch Recovery}\label{sec:sub_goal}


 {The Main Agent decomposes the instruction $\mathcal{T}$ into an ordered sequence of executable subgoals. Let $g_i$ denote the $i$-th subgoal in this sequence:}
\begin{equation}
    g_i =
    \left\langle
        \mathrm{Target}_i,\,
        \mathrm{Change}_i,\,
        \mathrm{Criterion}_i
    \right\rangle.
\end{equation}
 {where $\mathrm{Target}_i$ identifies the relevant entity, $\mathrm{Change}_i$ describes the intended state transition, and $\mathrm{Criterion}_i$ specifies the condition for verifying completion. The resulting subgoal sequence and its execution status are maintained in \texttt{PLAN.md}.}

 {After a subgoal is executed, itsexecuting a subgoal, the observed outcome is assessevaluated against theits completion criterion. If the criterion is satisfied, the Main Agent marks the subgoal as complete and proceeds to the next one. Otherwise, isubgoal is marked as complete and execution advances. When verification detects a failure, the Main Agent generates a textual diagnosis of the observed discrepancy, inspired by the verbal-reflection mechanism of Reflexion~\citep{shinn2023reflexion}. 
Based on this diagnosis, the Main Agent formulateit constructs a localized recovery subgoal and pushes it onto the Branch Stack. The stack-top recovery subgoal is executaddressed first; upon success, it is removed from the stackonce it is completed, the branch is removed and execution resumes from the interrupted plan. 
If a recovery subgoal continues to fail after a preset number of attempts~(\textit{e.g.}, two)fails , the Main Agent formulates an alternativea few times (\textit{e.g.}, twice) in a row, then  recovery subgoal and pushes it onto the Branch Stack.
the Main Agent revises the high-level plan.
}

\subsubsection{Token-Aware Memory Consolidation Engine}

 {EMERGE-Policy maintains four complementary forms of task state, together with a separate recovery stack. The active context window $\mathcal{C}_t$ serves as the Main Agent's working memory. \texttt{PLAN.md} stores the current subgoal sequence and completion status. \texttt{HISTORY.md} is an append-only chronological archive of summarized observations, actions, and outcomes. \texttt{MEMORY.md} is a consolidated belief state containing task-relevant spatial, environmental, and procedural facts; unlike the chronological history, it may be revised when later evidence changes an earlier belief. The Branch Stack separately stores transient recovery subgoals.}

 {Memory consolidation is triggered when the active context window reaches a  token capacity.}
 {A Memory Integration Agent then appends the evicted chronological summary to \texttt{HISTORY.md}, merges relevant facts into \texttt{MEMORY.md} while revising stale entries, and reconstructs the next active context $\mathcal{C}_{t+1}$ from the instruction, active \texttt{PLAN.md} state, Branch Stack, relevant consolidated facts, and recent Sub-Agent results. The full \texttt{HISTORY.md} archive is not automatically reinserted into the active context window. This operation changes only external textual state; it does not update model weights or Skill implementations.}


 {During each evaluation episode, \texttt{PLAN.md}, \texttt{HISTORY.md}, \texttt{MEMORY.md}, and the Branch Stack accumulate task state, summarized experience, consolidated facts, and recovery subgoals, respectively. All four structures are reinitialized before the next episode. Thus, no plan state, execution history, consolidated memory, or recovery state is transferred across episodes; the reported evaluation does not use cross-episode online adaptation, and episode order does not change the memory available to the agent.
}

\subsection{Sub Agents: Role-Specific Evidence Processing}
\label{sub agent}

 {Sub Agents perform evidence-intensive operations that support the Main Agent's task-level decisions. Each Sub Agent operates within a role-specific context and receives only the information required for its assigned request. This separation keeps raw observations, intermediate tool outputs, and detailed processing traces outside the Main Agent's active context window $\mathcal{C}_t$. 
The Main Agent receives only the structured evidence needed to continue planning, verification, or recovery.}

\subsubsection{Asynchronous Execution and Context Isolation}

 {To prevent evidence processing from blocking the Main Agent's planning cycle, Sub Agents run as non-blocking background workers. When the Main Agent delegates a request, a dedicated Sub Agent instance is created with an isolated prompt context containing the active subgoal, task-relevant spatial bounds, and permitted tool primitives. This design allows independent requests to be processed asynchronously while the Main Agent retains control of the task state.}
 {For a perception, verification, or monitoring request, let $k$ identify the Sub Agent role. Given the raw multimodal observation history $\mathcal{O}_{0:t}^{\mathrm{raw}}$ and active subgoal $g_i$, the corresponding Sub Agent produces structured evidence}
\begin{equation}
    \mathcal{Z}_t^{(k)}
    =
    f_k
    \left(
        \mathcal{O}_{0:t}^{\mathrm{raw}};\,
        g_i
    \right),
\end{equation}
 {where $f_k$ denotes the role-specific processing procedure and $\mathcal{Z}_t^{(k)}$ denotes its structured output. The semicolon indicates that observation processing is conditioned on the active subgoal rather than performed over the complete scene without task context.}

 {Once processing is complete, the Sub Agent returns $\mathcal{Z}_t^{(k)}$ to the Main Agent through a structured text output or an explicit state update. Its intermediate observations, prompts, and tool traces remain within the isolated context, preserving the separation between evidence processing and task-level reasoning.}

\subsubsection{Categorized Role-Specific Sub-Agent Modules}

Depending on operational requirements, Sub Agents are categorized into three specialized roles:




\paragraph{Perception Sub Agent.}
 {It extracts task-relevant scene evidence from the available visual and spatial observations. Depending on the request, it may invoke vision-language or spatial-grounding tools to identify objects, estimate their states, or characterize relevant spatial relationships. Its outputs contain the resulting scene evidence rather than the complete perceptual processing trace.}

\paragraph{Verification Sub Agents.}
 {This agent assesses conditions associated with an active subgoal. Before execution, it may evaluate whether the required preconditions are satisfied. After execution, it compares the observed outcome with $\mathrm{Criterion}_i$ and returns a completion judgment together with the supporting evidence.}

\paragraph{Monitor Sub Agents.}
 {The agent inspects the observations and execution feedback available during physical interaction. When it detects a deviation from the expected execution state, it returns an anomaly report to the Main Agent. The specific monitoring signals depend on the sensors and execution backend available in each deployment.}


\subsubsection{Failure Handover and Closed-Loop Feedback}

 {When monitoring or verification identifies a failure, the responsible Sub Agent returns a structured failure report containing the observed evidence, failure type, relevant state changes, and a likely cause when one can be inferred. The Main Agent uses this report to generate the textual failure diagnosis described in \Cref{Main Agent}.}

 {The Main Agent then determines whether to retry the current operation, construct a localized recovery subgoal, or revise the high-level plan. This separation keeps evidence processing within the appropriate Sub Agent while preserving recovery and planning decisions within the Main Agent. The resulting handover closes the loop between execution, observation, verification, and recovery.}

\subsection{Unified Skill and Tool Orchestration}
\label{skills and tools}


 {EMERGE-Policy exposes heterogeneous robot capabilities through a unified Skill interface. A Skill represents a callable capability with a defined function, input schema, and output contract, while a tool API provides the concrete interface through which that capability is accessed. Sub Agents may use these interfaces to process role-specific evidence, whereas the Main Agent selects and composes Skills according to the current plan and active context window.}

\subsubsection{Functional Skill Categorization}\label{sec:skill}

 {EMERGE-Policy categorizes Skills according to their function in the decision loop rather than their underlying model architecture. The same model family may therefore support different Skill types depending on how its output is used. The library contains three functional categories.}

\paragraph{Operational Skills.}
 {Operational Skills produce actions that are executed by the robot or simulation environment. They may be implemented by analytical motion primitives, trajectory controllers, VLA policies, or WAM-based control policies. Given a subgoal and the required spatial parameters, an Operational Skill produces an executable command or action sequence and returns its execution status.}

\paragraph{Imagination Skills.}
 {Imagination Skills predict possible outcomes without directly changing the environment. They may use world models or action-conditioned prediction models to estimate future observations, compare candidate actions, or identify potential execution failures before physical commitment. Their predictions are returned as structured evidence for subsequent decision-making.}

\paragraph{Evaluation Skills.}
 {Evaluation Skills assess feasibility, progress, or completion using available observations and system state. They may be implemented by vision-language models, geometric checks, or environment-specific state queries. In particular, an Evaluation Skill can compare the observed outcome of an Operational Skill with the active subgoal criterion $\mathrm{Criterion}_i$.}

 {These categories describe what a Skill does, rather than which agent or model implements it. A Skill exposes a specific callable capability, whereas a Sub Agent receives a task-specific request, may invoke one or more Skills, and returns structured evidence to the Main Agent. This functional organization allows learned action policies, world models, perception models, analytical controllers, and system tools to be accessed through a common invocation interface. New capability implementations can be registered through the same Skill schema without changing the orchestration protocol. For example, a value estimator can be incorporated as an Evaluation Skill.}

\subsubsection{Concrete Tool APIs and Multimodal Interfaces}

 {EMERGE-Policy exposes system operations and perceptual capabilities through concrete tool APIs that can be invoked by the Main Agent or authorized Sub Agents. We organize these interfaces into execution and workspace APIs, which support task orchestration and physical dispatch, and perceptual APIs, which provide task-relevant scene evidence.}

\paragraph{Main Agent Execution and Workspace APIs.}
 {The Agent Mode API switches the system between planning and physical execution. In execution mode, the Embodied Action API interfaces with the robot backend to invoke rule-based motion primitives or end-to-end VLA policies. The File System API provides access to structured \texttt{JSON} and \texttt{YAML} files, as well as the planning and memory files maintained during execution. The Shell API supports system operations and hardware diagnostic checks, including robot-connectivity checks. The Message API communicates execution status and results to the user interface.}

\paragraph{Perception Sub Agent APIs.}
 {Perception Sub Agents use the Scene Graph API and external vision interfaces to obtain task-relevant spatial evidence. These interfaces support object localization, image interpretation, and the generation of visual prompts for segmentation models such as SAM3. For tasks involving articulated or spatially constrained objects, such as opening doors, they may also estimate object relationships and viewpoint-dependent affordance maps. The resulting evidence is returned to the Main Agent to ground target selection and spatial constraints before VLA execution. The organization of these interfaces follows a hierarchical structure that separates execution dispatch from perceptual grounding, as illustrated in \Cref{fig:skills_and_tools}. This separation allows the Main Agent to invoke high-level execution APIs without being exposed to the complexity of perception pipelines, while Perception Sub Agents can operate on rich visual inputs without polluting the planning context. The standardized interface schema further ensures that new tools or backends can be integrated with minimal changes to the orchestration logic, preserving the extensibility of the framework across different robot platforms and task domains.}

\subsubsection{Standardized Schema and Dual-Mode Execution Lifecycle}

\begin{figure}
    \centering
    \includegraphics[width=0.75\linewidth]{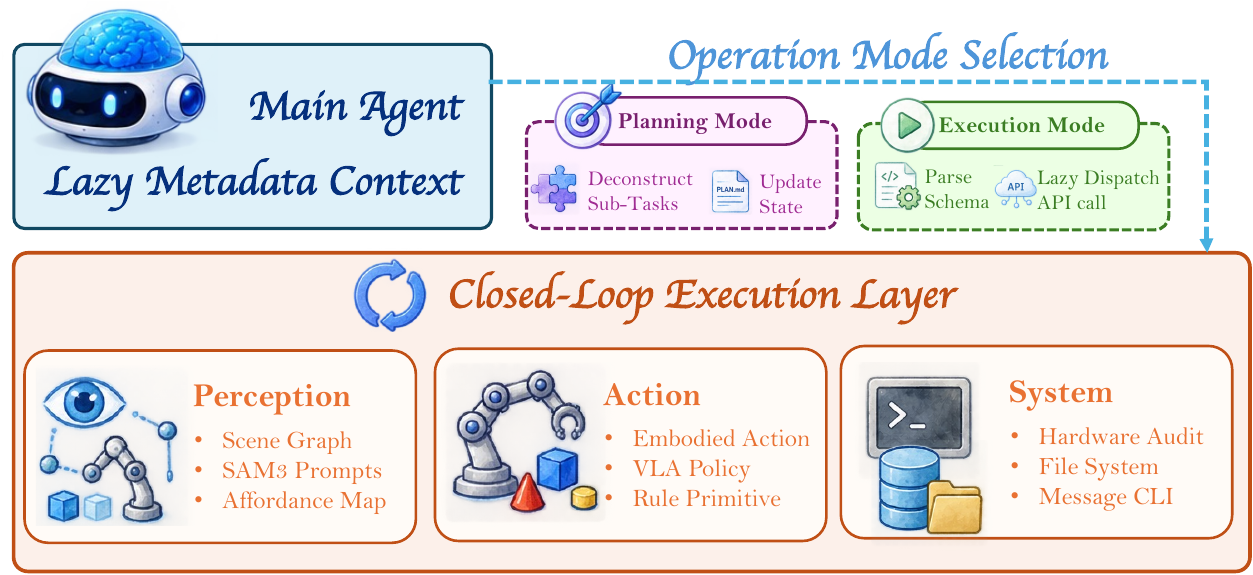}
    \caption{{\textbf{Tool API hierarchy and standardized interfaces.} EMERGE-Policy organizes callable interfaces into three groups: Perception APIs provide spatial and semantic evidence, Action APIs dispatch motion primitives or VLA policies, and System APIs support workspace access and hardware operations. Each interface follows a common schema, \( \langle \mathcal{I}_{\mathrm{meta}}, \Theta_{\mathrm{param}}, \Phi_{\mathrm{pre}}, \Psi_{\mathrm{post}} \rangle \), allowing the Main Agent to retain compact metadata and retrieve full parameters, preconditions, and postconditions only when the interface is invoked.}}
    \label{fig:skills_and_tools}
\end{figure}

 {As shown in \Cref{fig:skills_and_tools}, each registered system or perceptual interface is represented by a common functional schema}
\begin{equation}
    K =
    \left\langle
        \mathcal{I}_{\mathrm{meta}},
        \Theta_{\mathrm{param}},
        \Phi_{\mathrm{pre}},
        \Psi_{\mathrm{post}}
    \right\rangle,
\end{equation}
 {where $K$ denotes a callable interface, $\mathcal{I}_{\mathrm{meta}}$ is its single-line capability description, $\Theta_{\mathrm{param}}$ specifies its runtime parameters, $\Phi_{\mathrm{pre}}$ defines the conditions required for invocation, and $\Psi_{\mathrm{post}}$ defines the expected conditions used for result verification.}
 {For each available interface, the Main Agent initially includes only $\mathcal{I}_{\mathrm{meta}}$ in the active context window $\mathcal{C}_t$. Once an interface has been selected, its parameter specification, preconditions, and postconditions are retrieved for invocation. This lazy-loading mechanism limits the context occupied by interfaces that are not relevant to the current decision.}





 {Skill invocation follows a structured lifecycle governed by the Agent Mode framework.}

\paragraph{I. Planning Phase.}
 {In planning mode, the Main Agent decomposes the instruction $\mathcal{T}$ into an ordered sequence of subgoals $g_i$ and records their execution status in the episode-specific \texttt{PLAN.md} file. The File System API is used to retrieve the domain-specific directives and structured workspace information required for Skill selection and parameter construction. Imagination and Evaluation Skills may also be invoked during this phase to predict candidate outcomes or assess relevant constraints.}

\paragraph{II. Perceptual Grounding Phase.}
 {The Main Agent delegates perceptual grounding to a Perception Sub Agent. The Sub Agent queries the Scene Graph API and, when required, other perceptual interfaces to obtain object bounding boxes, spatial relationships, and affordance maps for candidate targets. The returned evidence grounds the target objects and spatial constraints used in subsequent planning.}

\paragraph{III. Physical Dispatch Phase.}
 {Once the spatial constraints and Skill preconditions are satisfied, the Agent Mode API transitions the system to execution mode. The Main Agent invokes the selected Operational Skill through the Embodied Action API, which dispatches either a rule-based trajectory primitive or a fine-tuned VLA policy, such as $\pi_{0.5}$. During execution, the Message API reports execution status, while the Shell API supports robot-connectivity checks.}

\paragraph{IV. Outcome Verification Phase.}
 {After physical execution, a Verification Sub Agent evaluates the resulting observation against the active subgoal criterion $\mathrm{Criterion}_i$. Successful verification updates \texttt{PLAN.md} and advances execution to the next subgoal. Failed verification produces a structured failure report that is returned to Main Agent, which then initiates localized Branch Stack recovery or revises the high-level plan.}
\section{Experiments}
\label{sec:Experiments}
 {We organize the evaluation around five research questions that test whether EMERGE-Policy provides system-level benefits beyond its execution backend and which design choices support those benefits:}
\begin{itemize}[leftmargin=*]
    \item  {\textbf{RQ1: System-level benchmark performance.} How does EMERGE-Policy compare with its underlying execution backends and representative agentic systems on standard and perturbed manipulation benchmarks?}
    \item  {\textbf{RQ2: Evidence for the Sub-Agent workflow.} Under matched tools, observations, and interaction budgets, what benefit is attributable to isolated Perception, Verification, Monitor, and Memory Sub Agents relative to a single-agent harness?}
    \item  {\textbf{RQ3: Action Selection and Failure Avoidance via Imagination.} How do Imagination and Evaluation Skills affect candidate-action selection, failure avoidance, and final task success?}
    \item  {\textbf{RQ4: Memory Architecture.} How does EMERGE-Policy manage long-horizon visual memory and environment state tracking to prevent error accumulation and information loss during complex manipulation tasks?}
    \item  {\textbf{RQ5: Physical deployment.} What success rate, execution cost, and failure profile does EMERGE-Policy exhibit on a multistep real-robot task?}
    \item  {\textbf{RQ6: Task-Specific Priors and Trajectory Optimization for Execution Efficiency.} How does incorporating task-specific planning priors into the Main Agent influence trajectory efficiency and overall operational overhead?}
\end{itemize}
\subsection{Experimental Setup}\label{sec:Experimental Setup}

We evaluate EMERGE-Policy on four benchmarks: LIBERO, LIBERO-Plus, and LIBERO-Pro for tabletop manipulation, and RoboDojo for bimanual manipulation.

Across all experiments, EMERGE-Policy uses the closed-source Codex prowered by codex-5.6-sol as the high-level planner and shares the same fixed set of basic action primitives across benchmarks and experimental settings. To accommodate differences in environment interfaces and task structures, we use a benchmark-specific \texttt{AGENTS.md} file to provide appropriate high-level instructions for the agent’s reasoning and interaction logic. These instructions guide how the agent operates within each benchmark, while the underlying planner and action primitive set remain unchanged. For fine-grained action execution, we consider two different primitive implementations: under the benchmark settings of the LIBERO series, WAM, instantiated with \texttt{Cosmos-Policy-LIBERO-Predict2-2B}\footnote{\url{https://huggingface.co/nvidia/Cosmos-Policy-LIBERO-Predict2-2B}}, and VLA, instantiated with \texttt{pi05\_LIBERO}\footnote{\url{https://github.com/Physical-Intelligence/openpi}}. Both policies use their officially released checkpoints without additional fine-tuning.

 {Cosmos Policy is exposed through two functional interfaces. As an Operational Skill, it proposes executable action chunks; as an Imagination Skill, it predicts the corresponding future frames. An Evaluation Skill scores the predicted outcomes, and greedy selection dispatches the highest-scoring candidate for execution.}

For LIBERO, LIBERO-Plus, and LIBERO-Pro, we augment the original evaluation environment with five additional fixed camera views to improve the agent's global scene perception and object localization. These additional cameras are used only by the agent-level perception and planning modules and are not included in the visual inputs to either WAM or VLA. WAM and VLA therefore retain their original observation interfaces, and the additional views do not alter the input space or execution behavior of the underlying low-level policies. The extra visual information is used exclusively for high-level perception and planning.

In addition, because EMERGE-Policy involves multi-stage perception, planning, and action execution, it typically requires longer interaction trajectories than direct policy execution. We therefore moderately increase the per-episode step limit relative to the original benchmark settings to prevent otherwise valid executions from being prematurely terminated due to an insufficient interaction budget.


\begin{table*}[t]
\centering
\small
\begin{tabular}{lccccc}
\toprule
\textbf{Method} & \textbf{Goal} & \textbf{Spatial} & \textbf{Object} & \textbf{Long} & \textbf{Average} \\
\midrule
\multicolumn{6}{c}{\textit{End-to-end policies and world-action models}} \\
\midrule
Diffusion Policy~\citep{chi2025diffusion} & 68.3 & 78.3 & 92.5 & 50.5 & 72.4 \\
Octo~\citep{team2024octo} & 84.6 & 78.9 & 85.7 & 51.1 & 75.1 \\
MDT~\citep{reuss2024multimodal} & 73.5 & 78.5 & 87.5 & 64.8 & 76.1 \\
OpenVLA~\citep{kim2024openvla} & 79.2 & 84.7 & 88.4 & 53.7 & 76.5 \\
$\pi_0$-FAST~\citep{pertsch2025fast} & 88.6 & 96.4 & 96.8 & 60.2 & 85.5 \\
$\pi_0$~\citep{black2024pi_0} & 95.8 & 96.8 & 98.8 & 85.2 & 94.2 \\
OpenVLA-OFT~\citep{kim2025fine} & 97.9 & 97.6 & 98.4 & 94.5 & 97.1 \\
$\pi_{0.5}$~\citep{intelligence2025pi_} & 98.0 & 98.8 & 98.2 & 92.4 & 96.8 \\
Fast-WAM~\citep{yuan2026fast} & 97.0 & 98.2 & \textbf{100} & 95.2 & 97.6 \\
StarVLA-$\alpha$~\citep{ye2026starvla} & \underline{98.6} & 98.7 & 99.7 & 94.2 & 97.8 \\
AtomicVLA~\citep{zhang2026atomicvla} & 97.2 & 98.8 & \underline{99.8} & 96.2 & 97.8 \\
Cosmos Policy~\citep{kim2026cosmos} & 98.2 & 98.1 & \textbf{100} & 97.6 & 98.5 \\
\midrule
\multicolumn{6}{c}{\textit{Harness-based frameworks}} \\
\midrule
HarnessVLA~\citep{zhang2026harness} & 94.0 & 97.0 & \textbf{100} & 93.0 & 96.0 \\
RoboHarness~\citep{huang2026roboharness} & -- & -- & -- & -- & 98.7 \\
\midrule
\multicolumn{6}{c}{\textbf{EMERGE-Policy}} \\
\midrule
EMERGE-Policy (w/o WM) & 98.2 \textbf{\textcolor{myred}{(+0.2)}} & \textbf{99.2} \textbf{\textcolor{myred}{(+0.4)}} & \textbf{100} \textbf{\textcolor{myred}{(+1.8)}} & \underline{97.8} \textbf{\textcolor{myred}{(+5.4)}} & \underline{98.8} \textbf{\textcolor{myred}{(+2.0)}} \\
EMERGE-Policy (w/ WM) & \textbf{99.2} \textbf{\textcolor{myred}{(+1.0)}} & \underline{99.0} \textbf{\textcolor{myred}{(+0.9)}} & \textbf{100} \textbf{\textcolor{myred}{(+0.0)}} & \textbf{98.6} \textbf{\textcolor{myred}{(+1.0)}} & \textbf{99.2} \textbf{\textcolor{myred}{(+0.7)}} \\

\bottomrule
\end{tabular}
\caption{ {{\textbf{Standard LIBERO results across four task suites.}} {Success rates (\%) are reported for LIBERO-Goal, LIBERO-Spatial, LIBERO-Object, and LIBERO-Long, with their mean.} 
{\textbf{Bold} and \underline{underlined} values denote the best and second-best results, respectively.} Here, WM denotes the world-model pathway: \emph{w/o WM} uses $\pi_{0.5}$ as the execution backend without this pathway, whereas \emph{w/ WM} enables the Cosmos Policy world-model pathway. {For EMERGE-Policy variants, red values in parentheses report absolute gains over the corresponding execution backend.}}}
\label{tab:LIBERO-four-tasks}
\end{table*}

\subsection{Results on Benchmarks}

\subsubsection{Baselines} 

We compare our approach with representative works of three types of methods: End-to-End VLAs and World-Action Models, Code-based and Harness-based Methods, on some mainstream benchmarks. The VLA and WAM baselines are: Baseline methods include Diffusion Policy~\citep{chi2025diffusion}, Octo~\citep{team2024octo}, MDT~\citep{reuss2024multimodal}, OpenVLA~\citep{kim2024openvla}, \(\pi_0\)-FAST~\citep{pertsch2025fast}, \(\pi_0\)~\citep{black2024pi_0}, OpenVLA-OFT~\citep{kim2025fine}, \(\pi_{0.5}\)~\citep{intelligence2025pi_}, Fast-WAM~\citep{yuan2026fast}, StarVLA-\(\alpha\)~\citep{ye2026starvla}, AtomicVLA~\citep{zhang2026atomicvla} and Cosmos Policy~\citep{kim2026cosmos}. For Code-based methods, We mainly compare CaP-X~\citep{fu2026cap}, RATS~\citep{zhang2026playful} and ASPIRE~\citep{lu2026aspire} because their performance has been published on the widely adopted benchmark. When it comes to the Harness-based robot policy, we mainly compare it with the two works HarnessVLA~\citep{zhang2026harness} RoboHarness~\citep{huang2026roboharness}.

\subsubsection{Results} 
In this section, we will present the results of our method on the benchmarks and answer the \textbf{RQ1} and \textbf{RQ2}.

\textbf{ {RQ1: System-level benchmark performance.}}  {On standard LIBERO (\Cref{tab:LIBERO-four-tasks}), EMERGE-Policy with the world-model pathway obtains a $99.2\%$ average success rate, compared with $98.5\%$ for its Cosmos Policy execution backend; the largest suite-level difference is on LIBERO-10 ($98.6\%$ versus $97.6\%$). The variant using $\pi_{0.5}$ obtains $98.8\%$, compared with $96.8\%$ for the underlying policy. Because performance is near the benchmark ceiling, these differences should be interpreted together with trial counts and uncertainty estimates rather than as evidence of a broad breakthrough.}

 {On LIBERO-Plus (\Cref{tab:LIBERO_plus}), the world-model variant obtains $93.9\%$, compared with $82.2\%$ for Cosmos Policy and $93.2\%$ for the reported RoboHarness result. These numbers indicate that the complete EMERGE-Policy system is robust under the evaluated perturbations. They do not, by themselves, identify which architectural component causes the gain, and comparisons with published baselines are controlled only when observation, camera, and interaction budgets match.}

\textbf{ {RQ2: Evidence for the Sub-Agent workflow.}}  {On LIBERO-Pro (\Cref{tab:LIBERO-pro-tasks}), EMERGE-Policy improves substantially over $\pi_{0.5}$ on several implicit-instruction suites, including Spat-S ($95.7\%$ versus $20.0\%$) and L10-S ($59.1\%$ versus $8.0\%$). This result is consistent with the intended benefits of high-level re-grounding and closed-loop orchestration. However, a comparison between the complete system and an end-to-end execution policy does not isolate the contribution of asynchronous cognitive decomposition. A matched ablation must compare the proposed Sub-Agent workflow with a single Main Agent that receives the same cameras, tools, prompts, and interaction budget.}


\begin{table*}[t]
\centering
\resizebox{\linewidth}{!}{%
\begin{tabular}{lcccccccc}
\toprule
\textbf{Method} & \textbf{State} & \textbf{Language} & \textbf{Layout} & \textbf{Background} & \textbf{Sensor} & \textbf{Camera} & \textbf{Light} & \textbf{Average} \\
\midrule
$\pi_0$~\citep{black2024pi_0} & 61.0 & 63.5 & 76.4 & 79.0 & 80.1 & 61.0 & 85.0 & 53.6 \\
$\pi_{0.5}$~\citep{intelligence2025pi_} & 75.4 & 85.6 & 85.7 & 94.6 & 89.7 & 75.4 & 96.9 & 85.7 \\
UniVLA~\citep{bu2025univla} & 46.2 & 77.6 & 31.9 & 81.0 & 21.2 & 1.8 & 69.0 & 42.9 \\
OpenVLA~\citep{kim2024openvla} & 3.5 & 23.0 & 28.5 & 34.8 & 15.2 & 0.8 & 8.1 & 15.6 \\
WorldVLA~\citep{cen2025worldvla} & 27.9 & 41.6 & 38.0 & 17.1 & 10.9 & 0.1 & 43.7 & 25.0 \\
RIPT-VLA~\citep{tan2025interactive} & 31.2 & 77.6 & 74.2 & 91.6 & 73.5 & 55.2 & 88.4 & 68.4 \\
DreamVLA~\citep{zhang2026dreamvla} & 17.6 & 67.0 & 43.5 & 71.5 & 53.6 & 26.2 & 77.5 & 48.9 \\
ABot-M0~\citep{yang2026abot} & 67.9 & 86.4 & 82.6 & 91.6 & 86.4 & 60.4 & 96.2 & 80.5 \\
Being-H0.7~\citep{luo2026being} & -- & -- & -- & -- & -- & -- & -- & 82.1 \\
OpenVLA-OFT~\citep{kim2025fine} & 21.7 & 81.0 & 68.7 & 91.0 & 78.6 & 55.6 & 92.7 & 67.9 \\
X-VLA~\citep{zheng2026x} & 89.7 & 75.7 & 71.8 & 96.0 & 62.7 & 23.4 & 88.2 & 71.4 \\
Cosmos-Policy~\citep{kim2026cosmos} & 63.3 & 81.7 & 82.2 & 88.9 & \underline{92.7} & 75.8 & 96.5 & 82.2 \\
Fast-WAM~\citep{yuan2026fast} & 44.5 & 68.9 & 60.7 & 53.7 & 37.7 & 16.4 & 78.2 & 51.5 \\
LingBot-VA~\citep{li2026causal} & 83.0 & 86.4 & 76.2 & 53.1 & 64.4 & 40.9 & 82.3 & 69.5 \\
RoboHarness~\citep{huang2026roboharness} & \underline{90.4} & \underline{97.0} & 86.8 & \textbf{97.1} & 90.3 & \textbf{87.6} & \underline{97.4} & \underline{93.2} \\
\midrule
\multicolumn{9}{c}{\textbf{EMERGE-Policy}} \\
\midrule
EMERGE-Policy (w/o WM) 
& \shortstack{87.6 \\ \textbf{\textcolor{myred}{(+12.2)}}} 
& \shortstack{87.5 \\ \textbf{\textcolor{myred}{(+1.9)}}} 
& \shortstack{\textbf{91.6} \\ \textbf{\textcolor{myred}{(+5.9)}}} 
& \shortstack{89.7 \\ \textbf{\textcolor{mygreen}{(-4.9)}}} 
& \shortstack{84.2 \\ \textbf{\textcolor{mygreen}{(-5.5)}}} 
& \shortstack{\underline{84.9} \\ \textbf{\textcolor{myred}{(+9.5)}}} 
& \shortstack{92.4 \\ \textbf{\textcolor{mygreen}{(-4.5)}}} 
& \shortstack{88.3 \\ \textbf{\textcolor{myred}{(+2.6)}}} \\
EMERGE-Policy (w/ WM) 
& \shortstack{\textbf{95.0} \\ \textbf{\textcolor{myred}{(+31.7)}}} 
& \shortstack{\textbf{98.0} \\ \textbf{\textcolor{myred}{(+16.3)}}} 
& \shortstack{\underline{89.0} \\ \textbf{\textcolor{myred}{(+6.8)}}} 
& \shortstack{\underline{97.0} \\ \textbf{\textcolor{myred}{(+8.1)}}} 
& \shortstack{\textbf{94.0} \\ \textbf{\textcolor{myred}{(+1.3)}}} 
& \shortstack{85.0 \\ \textbf{\textcolor{myred}{(+9.2)}}} 
& \shortstack{\textbf{100} \\ \textbf{\textcolor{myred}{(+3.5)}}} 
& \shortstack{\textbf{93.9} \\ \textbf{\textcolor{myred}{(+11.7)}}} \\
\bottomrule
\end{tabular}%
}
\caption{ {\textbf{LIBERO-Plus robustness under seven perturbation types.}} {Success rates (\%) are reported for each perturbation, with the mean in Avg. across all perturbations}. {\textbf{Bold} and \underline{underlined} values denote the best and second-best results, respectively.} For each EMERGE-Policy variant, colored values report the absolute change from its corresponding execution backend, $\pi_{0.5}$ or Cosmos Policy: \textcolor{myred}{\textbf{red}} indicates an improvement, and \textcolor{mygreen}{\textbf{green}} indicates a decrease.}
\label{tab:LIBERO_plus}
\end{table*}

\begin{table*}[t]
\centering
\resizebox{\linewidth}{!}{%
\begin{tabular}{lccccccccc}
\toprule
\textbf{Method} & \textbf{Spat-E} & \textbf{Spat-S} & \textbf{Obj-E} & \textbf{Obj-S} & \textbf{Goal-E} & \textbf{Goal-S} & \textbf{L10-E} & \textbf{L10-S} & \textbf{Total (E/S)} \\
\midrule
\multicolumn{10}{c}{\textbf{End-to-End VLAs}} \\
\midrule
$\pi_{0.5}$~\citep{intelligence2025pi_} & 46.0 & 20.0 & \underline{73.0} & 17.0 & 46.0 & 38.0 & 46.0 & 8.0 & 52.8 / 20.8 \\
OpenVLA~\citep{kim2024openvla} & \textbf{89.0} & 0.0 & 0.0 & 0.0 & \textbf{98.0} & 0.0 & \textbf{85.0} & 0.0 & \underline{68.0} / 0.0 \\
X-VLA~\citep{zheng2026x} & -- & 0.0 & -- & 2.0 & -- & 1.0 & -- & 0.0 & -- / 0.8 \\
$\pi_0$~\citep{black2024pi_0} & 60.0 & 0.0 & 29.0 & 0.0 & 39.0 & 0.0 & 27.0 & 0.0 & 38.8 / 0.0 \\
Molmoact~\citep{lee2025molmoact} & -- & 0.0 & -- & 6.0 & -- & 0.0 & -- & 0.0 & -- / 1.5 \\
$\pi_{\text{RLinf}}$ & -- & 59.0 & -- & 78.0 & -- & 42.0 & -- & 14.0 & -- / 48.3 \\
\midrule
\multicolumn{10}{c}{\textbf{Code-based and harness-based methods}} \\
\midrule
CaP-X~\citep{fu2026cap} & -- & 12.0 & -- & 22.0 & -- & 26.0 & -- & -- & -- / 20.0 \\
RATS~\citep{zhang2026playful} & -- & 29.0 & -- & 61.0 & -- & 43.0 & -- & -- & -- / 44.3 \\
HarnessVLA~\citep{zhang2026harness} 
& -- 
& \shortstack{\underline{69.0} \\ \textbf{\textcolor{myred}{(+10.0)}}} 
& -- 
& \shortstack{\underline{91.0} \\ \textbf{\textcolor{myred}{(+13.0)}}} 
& -- 
& \shortstack{\textbf{66.0} \\ \textbf{\textcolor{myred}{(+24.0)}}} 
& -- 
& \shortstack{\underline{49.0} \\ \textbf{\textcolor{myred}{(+35.0)}}} 
& \shortstack{-- / \underline{68.8} \\ \textbf{\textcolor{myred}{(--/+20.5)}}} \\
\midrule
\multicolumn{10}{c}{\textbf{EMERGE-Policy}} \\
\midrule
EMERGE-Policy 
& \shortstack{\underline{71.4} \\ \textbf{\textcolor{myred}{(+25.4)}}} 
& \shortstack{\textbf{95.7} \\ \textbf{\textcolor{myred}{(+75.7)}}} 
& \shortstack{\textbf{95.5} \\ \textbf{\textcolor{myred}{(+22.5)}}} 
& \shortstack{\textbf{100} \\ \textbf{\textcolor{myred}{(+83.0)}}} 
& \shortstack{\underline{56.5} \\ \textbf{\textcolor{myred}{(+10.5)}}} 
& \shortstack{\underline{56.5} \\ \textbf{\textcolor{myred}{(+18.5)}}} 
& \shortstack{\underline{65.4} \\ \textbf{\textcolor{myred}{(+19.4)}}} 
& \shortstack{\textbf{59.1} \\ \textbf{\textcolor{myred}{(+51.1)}}} 
& \shortstack{\textbf{72.2} / \textbf{77.8} \\ \textbf{\textcolor{myred}{(+19.4 / +57.0)}}} \\
\bottomrule
\end{tabular}%
}
\caption{ {{\textbf{LIBERO-Pro instruction robustness under explicit and implicit instructions.}} Success rates (\%) on four task suites under explicit ($E$) and implicit ($S$) instructions. {\textbf{Bold} and \underline{underlined} values denote the best and second-best results, respectively.} Colored values report absolute changes from each method's stated execution baseline, $\pi_{\mathrm{RLinf}}$ for HarnessVLA and $\pi_{0.5}$ for EMERGE-Policy: \textcolor{myred}{\textbf{red}} indicates an improvement, and \textcolor{mygreen}{\textbf{green}} indicates a decrease.}}
\label{tab:LIBERO-pro-tasks}
\end{table*}

\subsection{Real-Robot Deployment}

In this section, we will focus on the real robot experiments of EMERGE-Policy and analyze the robustness of the framework, as well as answer \textbf{RQ5}.

In this section, we mainly conducted a real-device experiment called \textbf{"Multi-layered Cup Stacking"}. The task of the experiment was to remove the paper cups from their designated positions, then place the cups upside down at specific locations so that the cups were close to each other in the same layer. After all the cups in the same layer were placed, more cups were stacked on top of them to form 1, 2, and 3 layers with 3, 2, and 1 cups respectively.

\begin{figure*}[!t]
    \centering
    \begin{minipage}[t]{0.68\textwidth}
        \vspace{0pt}
        \centering
        \subcaptionbox{{EMERGE-Policy workflow}\label{fig:wm-workflow}}{%
            \includegraphics[width=\linewidth]{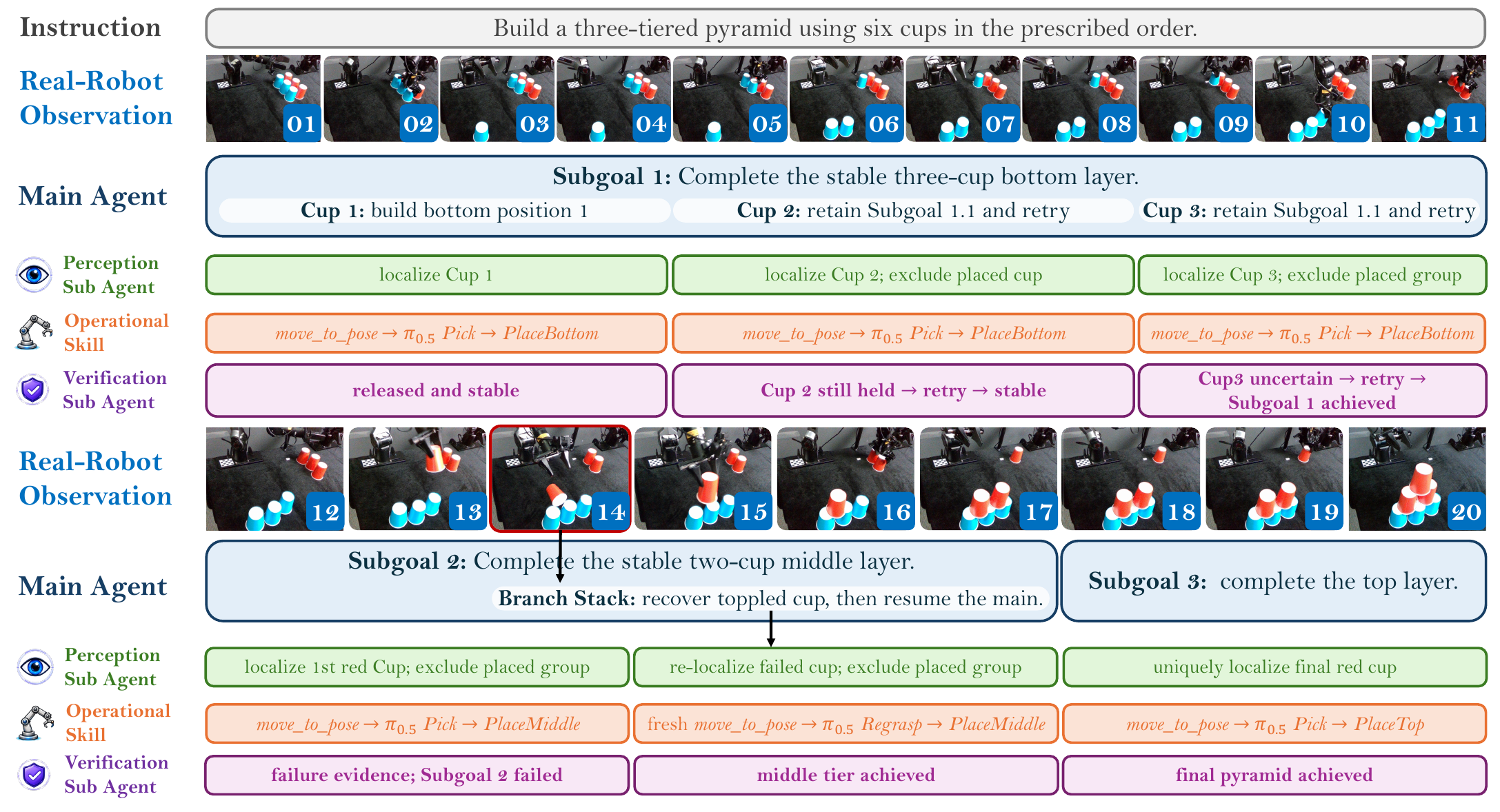}%
        }
    \end{minipage}\hfill
    \begin{minipage}[t]{0.30\textwidth}
        \vspace{0pt}
        \centering
        \subcaptionbox{Standard results\label{fig:real-robot-standard}}{%
            \includegraphics[width=\linewidth]{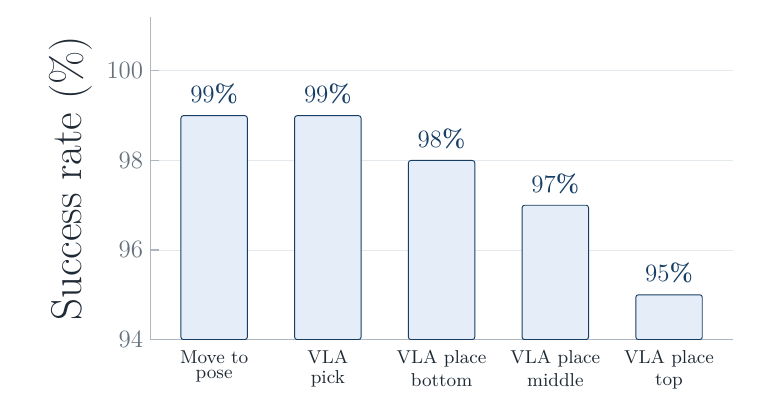}%
        }
        \vspace{0.8em}
        \subcaptionbox{Disturbance robustness\label{fig:real-robot-perturbation}}{%
            \includegraphics[width=\linewidth]{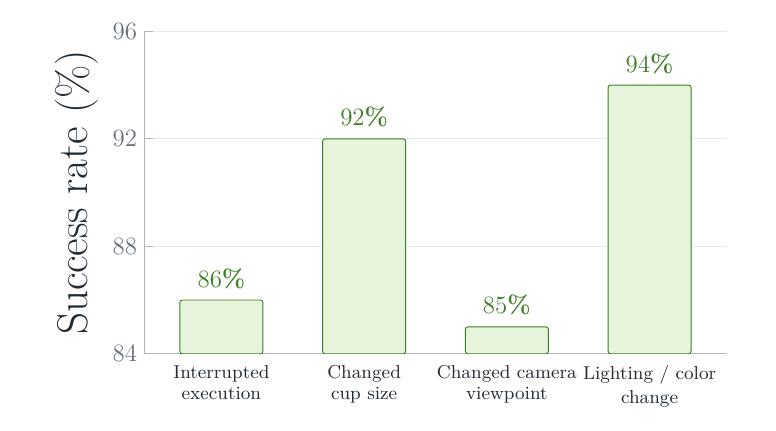}%
        }
    \end{minipage}
    \caption{{\textbf{Real-robot cup-stacking workflow and evaluation.} (a) The workflow traces Main Agent planning, Sub Agent perception and verification, and Operational Skill execution while constructing a three-tier cup pyramid. (b) Completion rates for the standard task stages. (c) Robustness under execution interruption, cup-size variation, camera changes, and lighting or color changes.}}
    \label{fig:experiments-overview}
\end{figure*}

Through experimental design and analysis, we found that this case can effectively measure the performance of each aspect of the framework: 
\begin{enumerate}[leftmargin=*]
    \item The entire task is a long-horizon task, and long-horizon tasks are an important criterion for evaluating the performance of embodied operation models. This task tests the collaboration of concurrent agents, the initiation and termination of skills, as well as the robustness and accuracy of actuator operations by the EMERGE-Policy when performing the task. 
    \item This task ingeniously integrates multiple performance tests. Firstly, the paper cups need to be precisely placed in a certain spatial position, which tests the positioning ability of the system. Secondly, in this task, the grippers need to perform multiple grasping and placement operations, which tests the functionality of the actuators of the system. Third, for Long-horizon tasks require the EMERGE-Policy to maintain the memory of its own progress, and the main agent needs to obtain information and decompose sub-tasks.
\end{enumerate}
We conducted 100 real robot experiments: we calculated the completion rates of five types of tasks from these 100 experiments, and conducted 30 experiments for each of the four interference settings. In each test construction, we randomly placed paper cups on the table and required the EMERGE-Policy to identify and pick up these paper cups, then transfer control to the planning policy to move to the designated position, and then let VLA stack them. The overall closed-loop workflow of the real-robot experiments is illustrated in \Cref{fig:wm-workflow}, which traces the sequential interaction among the Main Agent, Sub Agents, and Skills from task initialization through perception, physical execution, and visual verification. As shown in \Cref{fig:real-robot-standard}, the EMERGE-Policy achieved a high completion rate and stable performance. We also evaluated the four types of interference in the paper cup stacking task in \Cref{fig:real-robot-perturbation}. Changing the camera perspective and quantity resulted in the greatest performance decline, with the success rate dropping from 94\% to 85\%, because the reduction in the number of cameras affected the positioning tools and placement position calculations of the EMERGE-Policy. The EMERGE-Policy maintained stability. After interrupting the execution and partially completed structures were dismantled, the success rate reached 86\%, indicating the ability to respond online and re-plan. With a 5\% - 15\% change in the size of the paper cups, a success rate of 92\% could still be achieved, showing its tolerance for moderate perceptual errors. The influence of scene color change was very small, and the success rate remained at 93\%-94\%, indicating the ability to effectively identify and locate objects related to the task when the environment and desktop colors change.


\subsection{Ablation Study}


\begin{table*}[t]
\centering
\resizebox{\linewidth}{!}{%
\begin{tabular}{lcccccccc}
\toprule
\textbf{Method} & \textbf{State} & \textbf{Language} & \textbf{Layout} & \textbf{Background} & \textbf{Sensor} & \textbf{Camera} & \textbf{Light} & \textbf{Average} \\
\midrule
EMERGE-Policy 
& \textbf{95.0} 
& \textbf{98.0} 
& \underline{89.0} 
& \textbf{97.0} 
& \textbf{94.0} 
& \textbf{85.0} 
& \textbf{100} 
& \textbf{93.9} \\
\midrule
EMERGE-Policy (w/o Eval) 
& \shortstack{81.6 \\ \textbf{\textcolor{mygreen}{(-13.4)}}} 
& \shortstack{87.1 \\ \textbf{\textcolor{mygreen}{(-10.9)}}} 
& \shortstack{88.2 \\ \textbf{\textcolor{mygreen}{(-0.8)}}} 
& \shortstack{88.2 \\ \textbf{\textcolor{mygreen}{(-8.8)}}} 
& \shortstack{82.6 \\ \textbf{\textcolor{mygreen}{(-11.4)}}} 
& \shortstack{81.2 \\ \textbf{\textcolor{mygreen}{(-3.8)}}} 
& \shortstack{90.6 \\ \textbf{\textcolor{mygreen}{(-9.4)}}} 
& \shortstack{85.6 \\ \textbf{\textcolor{mygreen}{(-8.3)}}} \\

EMERGE-Policy (w/o Sub) 
& \shortstack{\underline{89.8} \\ \textbf{\textcolor{mygreen}{(-5.2)}}} 
& \shortstack{\underline{90.2} \\ \textbf{\textcolor{mygreen}{(-7.8)}}} 
& \shortstack{88.8 \\ \textbf{\textcolor{mygreen}{(-0.2)}}} 
& \shortstack{\underline{92.1} \\ \textbf{\textcolor{mygreen}{(-4.9)}}} 
& \shortstack{\underline{90.7} \\ \textbf{\textcolor{mygreen}{(-3.3)}}} 
& \shortstack{\underline{82.8} \\ \textbf{\textcolor{mygreen}{(-2.2)}}} 
& \shortstack{\underline{95.7} \\ \textbf{\textcolor{mygreen}{(-4.3)}}} 
& \shortstack{\underline{91.2} \\ \textbf{\textcolor{mygreen}{(-2.7)}}} \\

EMERGE-Policy (w/o WM) 
& \shortstack{87.6 \\ \textbf{\textcolor{mygreen}{(-7.4)}}} 
& \shortstack{87.5 \\ \textbf{\textcolor{mygreen}{(-10.5)}}} 
& \shortstack{\textbf{91.6} \\ \textbf{\textcolor{myred}{(+2.6)}}} 
& \shortstack{89.7 \\ \textbf{\textcolor{mygreen}{(-7.3)}}} 
& \shortstack{84.2 \\ \textbf{\textcolor{mygreen}{(-9.8)}}} 
& \shortstack{84.9 \\ \textbf{\textcolor{mygreen}{(-0.1)}}} 
& \shortstack{92.4 \\ \textbf{\textcolor{mygreen}{(-7.6)}}} 
& \shortstack{88.3 \\ \textbf{\textcolor{mygreen}{(-5.6)}}} \\
\bottomrule
\end{tabular}%
}
\caption{{\textbf{Ablation study on LIBERO-Plus.}} {Success rates (\%) are reported for each perturbation, with the mean in Avg. across all perturbations}. {\textbf{Bold} and \underline{underlined} values denote the best and second-best results, respectively.} Colored values report the absolute change relative to the full EMERGE-Policy baseline: \textcolor{myred}{\textbf{red}} indicates an improvement, and \textcolor{mygreen}{\textbf{green}} indicates a decrease.}
\label{tab:LIBERO_plus_ablation}
\end{table*}

In this section, we analyze the effect of each module in the task of libero-plus by shielding the different modules of the complete Emerge-policy. As shown in the ~\Cref{tab:LIBERO_plus_ablation}. Among them, EMERGE-Policy (w/o WM) indicates the elimination of the imagination skill of the world model during task execution, EMERGE-Policy (w/o Sub) represents the removal of the Verification Sub-Agent and Monitor Sub-Agent, and EMERGE-Policy (w/o Eval) indicates the elimination of the Evaluation skill of the Main-Agent. \Cref{tab:LIBERO_plus_ablation} presents a systematic breakdown of system performance across seven environmental perturbation modalities in LIBERO-Plus upon selectively deactivating key components of EMERGE-Policy. Removing the Evaluation Skill (w/o Eval) leads to the most severe overall degradation, dropping the average success rate from 93.9\% to 85.6\%—an absolute decline of 8.3\%. The performance drop is particularly pronounced under State (-13.4\%), Sensor (-11.4\%), and Language (-10.9\%) variations, demonstrating that real-time evaluation against subgoal criteria is vital when physical or perceptual states fluctuate unpredictably. Disabling the World Model's Imagination Skill (w/o WM) results in the second-largest overall drop, reducing average success to 88.3\% (-5.6\%), with steep declines in Language (-10.5\%) and Sensor (-9.8\%) robustness. Interestingly, removing the World Model leads to a slight performance gain in Layout (+2.6\%), suggesting that hallucination or over-reliance on dynamic imagination under static spatial shifts can occasionally introduce noise. Finally, disabling the Verification and Monitor Sub-Agents (w/o Sub) yields the second-best performance across most categories (91.2\% average, -2.7\%), but still causes notable drops under Language (-7.8\%) and State (-5.2\%) shifts. Overall, these results confirm that while context-isolated sub-agents contribute to baseline stability, the real-time Evaluation Skill and World Model imagination serve as the primary drivers of policy robustness.

\begin{figure}
    \centering
    \includegraphics[width=1.0\linewidth]{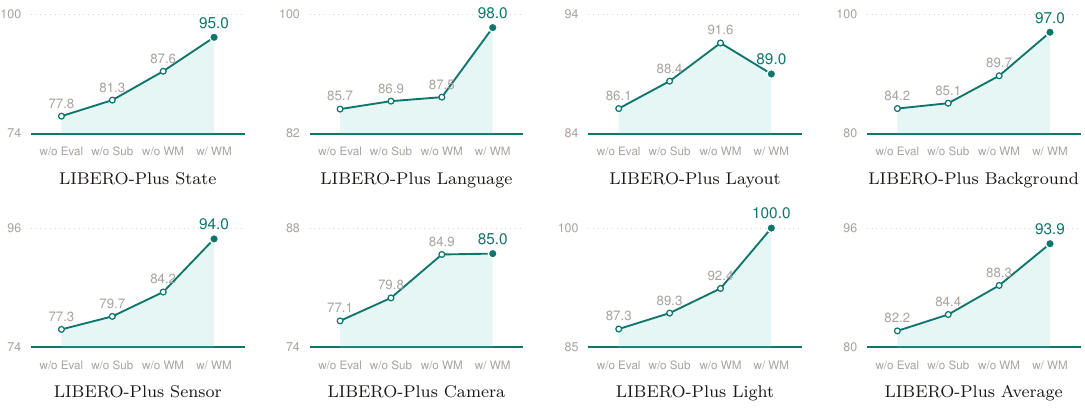}
    \caption{\textbf{Ablation study for performance emerging.} We successively removed different skills and sub-agents, and then observed the performance of the framework with various configuration settings on libero-plus.}
    \label{ablation_fig}
\end{figure}

Similarly, we also analyzed the changes in the performance of the entire framework when multiple modules were removed simultaneously. As shown in the ~\Cref{ablation_fig}, where "w/o WM" represents removing only the world model module, and "w/o Sub" represents removing the world model, Verification Sub-Agent and Monitor Sub-Agent. "w/o Eval" represents the removal of the "world model", "Sub-Agents" and "Evaluation skill". \Cref{ablation_fig} illustrates the cumulative performance degradation on LIBERO-Plus as skills and sub-agents are sequentially stripped away, providing strong empirical evidence for the emergent nature of EMERGE-Policy. Across individual perturbation dimensions—such as State, Language, Sensor, and Camera—performance exhibits a monotonic decrease as the system shifts from the full configuration toward progressively bare-bones variants. In the overall average metric (LIBERO-Plus Average), the success rate declines from 93.9\% in the complete framework down to 88.3\% when removing the World Model (w/o WM), 84.4\% when additionally stripping out Sub-Agents (w/o Sub), and hits a low of 82.2\% when the Evaluation Skill is also removed (w/o Eval). The non-linear compound drop across complex perturbations (e.g., dropping to 77.8\% on State and 77.3\% on Sensor in the fully stripped setup) demonstrates that high success rates do not originate from any single isolated backend. Instead, robust embodied intelligence explicitly emerges from the multi-agent orchestration graph, where task planning, context-isolated verification, world-model imagination, and criterion-grounded evaluation operate in closed-loop synergy.

\subsection{Experiment Case Analysis}

\subsubsection{Memory and Recovery}


\begin{figure}[ht]
    \centering
    \includegraphics[width=0.85\textwidth]{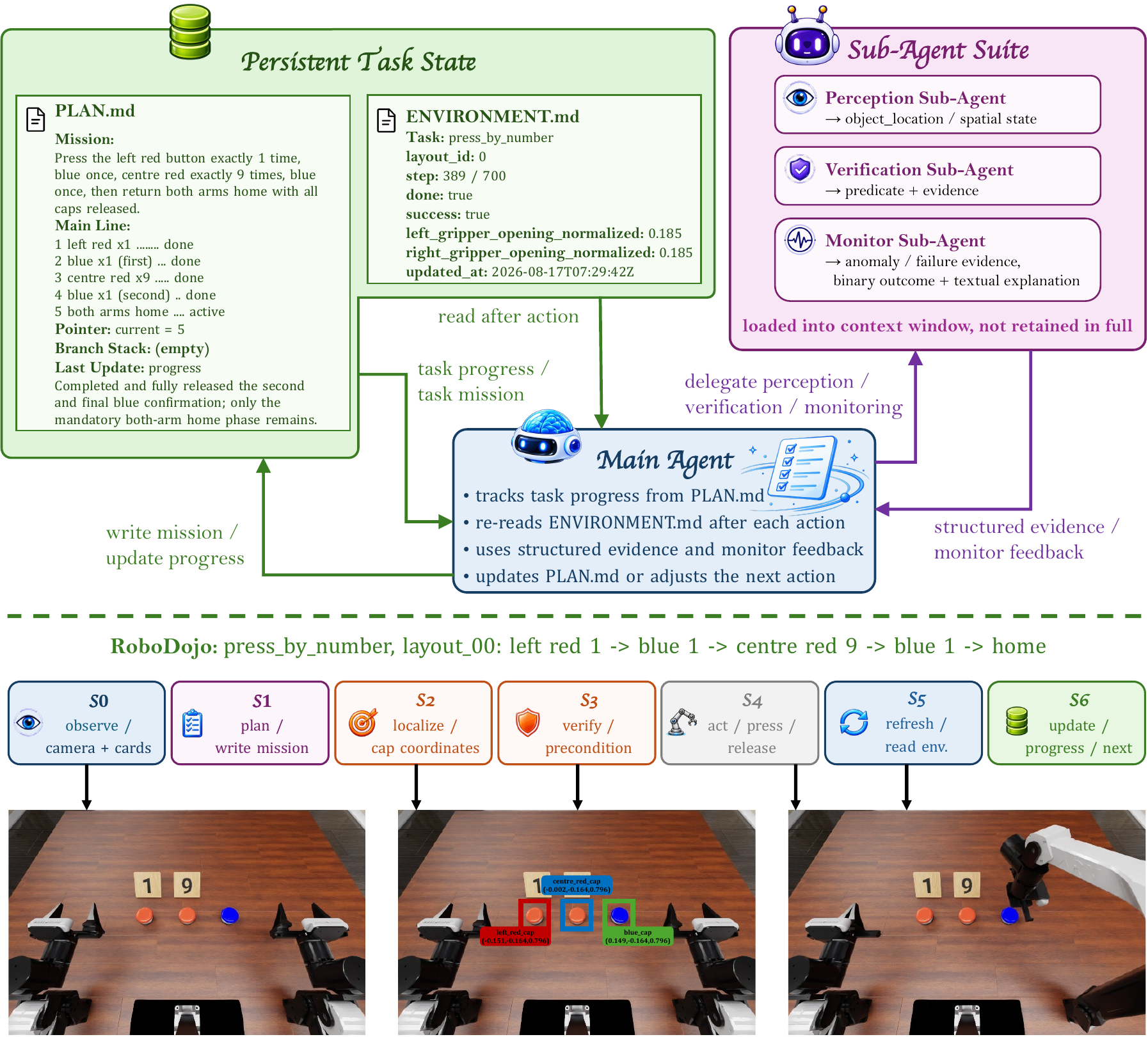}
    \caption{{\textbf{Three-tier memory and context management.} \texttt{PLAN.md} and \texttt{ENVIRONMENT.md} maintain episode task state and current environment feedback. Specialized Sub Agents convert dense observations into structured evidence for the Main Agent, which consolidates relevant facts and updates execution using asynchronous visual verification.}}
    \label{fig:memory}
\end{figure}

Next, we will answer \textbf{RQ4} by analyzing EMERGE-Policy. The memory architecture of EMERGE-Policy is not a monolithic neural state but a multi‑layer, externalised framework that decouples task progression from transient working memory, as illustrated in \Cref{fig:memory}. At its foundation, the agent maintains two persistent files that serve as the ground truth for long‑horizon execution. \texttt{PLAN.md} stores the complete subgoal decomposition along with the completion status of each step, and is updated through two distinct modes -- \textit{new mission} to initialise the overall objective and progress to incrementally record milestones such as finishing a press cycle or advancing to the confirmation phase. Complementing this, \texttt{ENVIRONMENT.md} is automatically refreshed by the environment watchdog after every physical action; it contains simulation‑level facts including joint states, step counters, and success flags. The agent mandatorily re‑reads this file after each \textit{execute robot action} call, ensuring that every subsequent decision is grounded in the latest physical snapshot rather than in stale context‑window contents. Together, these external state files provide a durable and inspectable record that enables safe recovery from interruptions and prevents the accumulation of outdated information within the limited token budget of the language model.

Above this persistent layer, the agent leverages structured evidence returned by specialised sub‑agents for perception, planning, and verification. Sub‑agents such as object location, task planning, and task verification produce compact JSON payloads containing explicitly labelled fields—for instance, spatial coordinates, predicate truth values with supporting evidence, or a step‑wise action list. These outputs are consumed by the main agent as short‑term working memory within its active context, but they are never retained in their full raw form. Instead, they are semantically distilled to retain only the decision‑relevant facts, such as the exact 3D position of each button cap or the binary outcome of a pre‑condition check. This design avoids cluttering the reasoning window with voluminous sensor traces or interaction histories, while still providing the precise, machine‑readable evidence required for subsequent planning and verification steps.

The third layer consists of a closed‑loop visual verification process that actively corrects the agent's remembered state. A dedicated visual monitor thread asynchronously evaluates the scene after each action, comparing observed visual cues against the expected subgoal criterion (e.g., whether the gripper is clear of the button). The monitor returns a binary judgement along with a textual explanation, which the main agent immediately incorporates into its updated belief. This feedback loop is critical because it catches discrepancies that may not be reflected in the motion execution result alone—for example, even when a recover command reports success, the visual monitor may indicate that the arm is still obstructing the view of the button, prompting further corrective actions. Consequently, the memory state is dynamically refined by perceptual evidence, rather than relying solely on proprioceptive or command‑level feedback.

Overall, this three‑tiered memory mechanism—persistent files for task and environment state, structured sub‑agent evidence for concise contextual facts, and asynchronous visual verification for error correction—collectively ensures that the main agent operates within a bounded context window while maintaining reliable awareness of progress and physical conditions. The separation of roles not only enhances scalability for long‑horizon tasks but also introduces transparency and auditability, as every major state transition and corrective decision is recorded in the external files or in the structured outputs of the sub‑agents. This architectural choice underpins the agent's robustness in perturbed scenarios and its ability to resume or recover from failures without relying on an unbounded internal memory, thus aligning with the system‑level orchestration philosophy of EMERGE‑Policy. Through these three memory mechanisms, we achieved the best performance in the memory and open settings of Robodojo-Sim, shown in \Cref{tab:robodojo_sim}, indicating that it possesses strong historical information encoding and non-Markovian reasoning capabilities, and it performed exceptionally well in skill reorganization and the alignment of open vocabulary semantics to actions. It can remember the key visual information that has been obscured or disappeared and use it for subsequent decisions. This three-tiered memory mechanism is also visually reflected in \Cref{fig:memory}, where the agent repeatedly integrates persistent files, structured sub-agent evidence, and visual verification feedback throughout execution.

\begin{table*}[t]
\centering
\resizebox{\linewidth}{!}{%
\begin{tabular}{lcccccc}
\toprule
\textbf{Method} & \textbf{Generalization} & \textbf{Memory} & \textbf{Precision} & \textbf{Long-Horizon} & \textbf{Open} & \textbf{Average} \\
\midrule
GalaxeaVLA (G0.5)~\citep{liu2026g05} & 18.95 / 13.00\% & 8.61 / 7.33\% & 28.25 / 20.42\% & 44.12 / 32.25\% & 1.73 / 1.58\% & 20.23 / 14.88\% \\
Xiaomi-Robotics-1~\citep{team2026xiaomi} & 23.55 / 17.00\% & 7.81 / 6.56\% & 26.69 / 18.83\% & 38.39 / 23.67\% & 3.94 / 3.58\% & 20.07 / 13.93\% \\
\rowcolor{gray!20}
EMERGE-Policy & 15.37 / 9.17\% & \textbf{25.00 / 22.51\%} & 12.40 / 7.38\% & 29.54 / 17.42\% & \textbf{19.98 / 11.20\%} & 20.46 / 13.54\% \\
Hy-Embodied-0.5-VLA~\citep{zhang2026hy} & 11.78 / 8.50\% & 13.37 / 12.11\% & 13.81 / 8.00\% & 25.74 / 14.92\% & 0.65 / 0.58\% & 13.07 / 8.80\% \\
Spatial Forcing~\citep{li2026spatial} & 14.12 / 9.50\% & 5.43 / 4.11\% & 17.33 / 10.58\% & 23.26 / 14.58\% & 1.78 / 1.58\% & 12.38 / 8.04\% \\
\rowcolor{gray!20}
$\pi_{0.5}$~\citep{intelligence2025pi_} & 13.38 / 8.00\% & 5.78 / 4.56\% & 12.40 / 5.50\% & 23.54 / 14.67\% & 1.98 / 1.67\% & 11.41 / 6.91\% \\
InternVLA-A1.5~\citep{ma2026internvla} & 10.36 / 7.00\% & 4.93 / 3.56\% & 15.23 / 10.17\% & 23.80 / 13.75\% & 1.43 / 1.42\% & 11.15 / 7.14\% \\
VLAct~\citep{yang2026vlactmodel} & 9.54 / 6.50\% & 0.66 / 0.56\% & 20.62 / 15.25\% & 20.12 / 13.67\% & 2.37 / 2.25\% & 10.66 / 7.60\% \\
X-VLA~\citep{zheng2026x} & 10.47 / 6.50\% & 4.76 / 3.56\% & 18.32 / 12.00\% & 16.53 / 9.75\% & 0.55 / 0.50\% & 10.13 / 6.52\% \\
X-WAM~\citep{guo2026unified} & 7.39 / 3.00\% & 6.32 / 4.67\% & 6.72 / 1.83\% & 17.47 / 9.08\% & 0.57 / 0.25\% & 7.69 / 3.83\% \\
Xiaomi-Robotics-0~\citep{cai2026xiaomi} & 7.43 / 5.50\% & 5.07 / 3.67\% & 8.42 / 4.58\% & 13.51 / 6.92\% & 0.22 / 0.17\% & 6.93 / 4.18\% \\
StarVLA~\citep{ye2026starvla} & 3.94 / 2.50\% & 3.34 / 2.44\% & 9.90 / 4.33\% & 14.15 / 6.50\% & 0.68 / 0.58\% & 6.40 / 3.24\% \\
GigaWorld-Policy-0~\citep{ye2026gigaworld} & 5.35 / 3.00\% & 3.46 / 2.22\% & 6.15 / 1.83\% & 15.51 / 8.92\% & 0.54 / 0.50\% & 6.20 / 3.27\% \\
GalaxeaVLA (G0)~\citep{jiang2025galaxea} & 4.54 / 3.00\% & 3.17 / 1.89\% & 8.10 / 3.83\% & 12.60 / 5.58\% & 0.70 / 0.67\% & 5.82 / 2.96\% \\
LingBot-VLA~\citep{wu2026pragmatic} & 6.72 / 4.50\% & 3.82 / 2.78\% & 5.33 / 1.83\% & 10.89 / 5.25\% & 0.72 / 0.67\% & 5.50 / 2.96\% \\
EventVLA~\citep{yang2026eventvla} & 3.95 / 1.50\% & 4.92 / 4.78\% & 10.13 / 5.75\% & 5.05 / 0.83\% & 0.80 / 0.75\% & 4.97 / 2.81\% \\
AHA-WAM~\citep{cai2026aha} & 5.79 / 3.00\% & 2.97 / 2.78\% & 5.86 / 2.42\% & 8.61 / 2.67\% & 0.88 / 0.83\% & 4.82 / 2.39\% \\
ABot-M0~\citep{yang2026abot} & 5.73 / 3.50\% & 3.96 / 0.50\% & 2.44 / 2.22\% & 5.50 / 1.75\% & 0.72 / 0.67\% & 3.67 / 1.73\% \\
\midrule
\multicolumn{7}{c}{\textbf{Human Expert (Teleoperation)}} \\
\midrule
Human Expert~\citep{robodojo2026} & -- & -- & -- & -- & -- & 80.42 / 76.03\% \\
\bottomrule
\end{tabular}%
}
\caption{ {{\textbf{RoboDojo simulation results across five capability dimensions.}} Each entry reports the score and success rate (\%) for five capability dimensions and their average. Published baseline and human-expert results are reproduced from RoboDojo~\citep{robodojo2026} and its public leaderboard; EMERGE-Policy is evaluated using the same protocol. An em dash indicates that a result is not reported.}}
\label{tab:robodojo_sim}
\end{table*}

\subsubsection{Imagination and Evaluation Skills}

\begin{figure}[ht]
    \centering
    \includegraphics[width=0.7\textwidth]{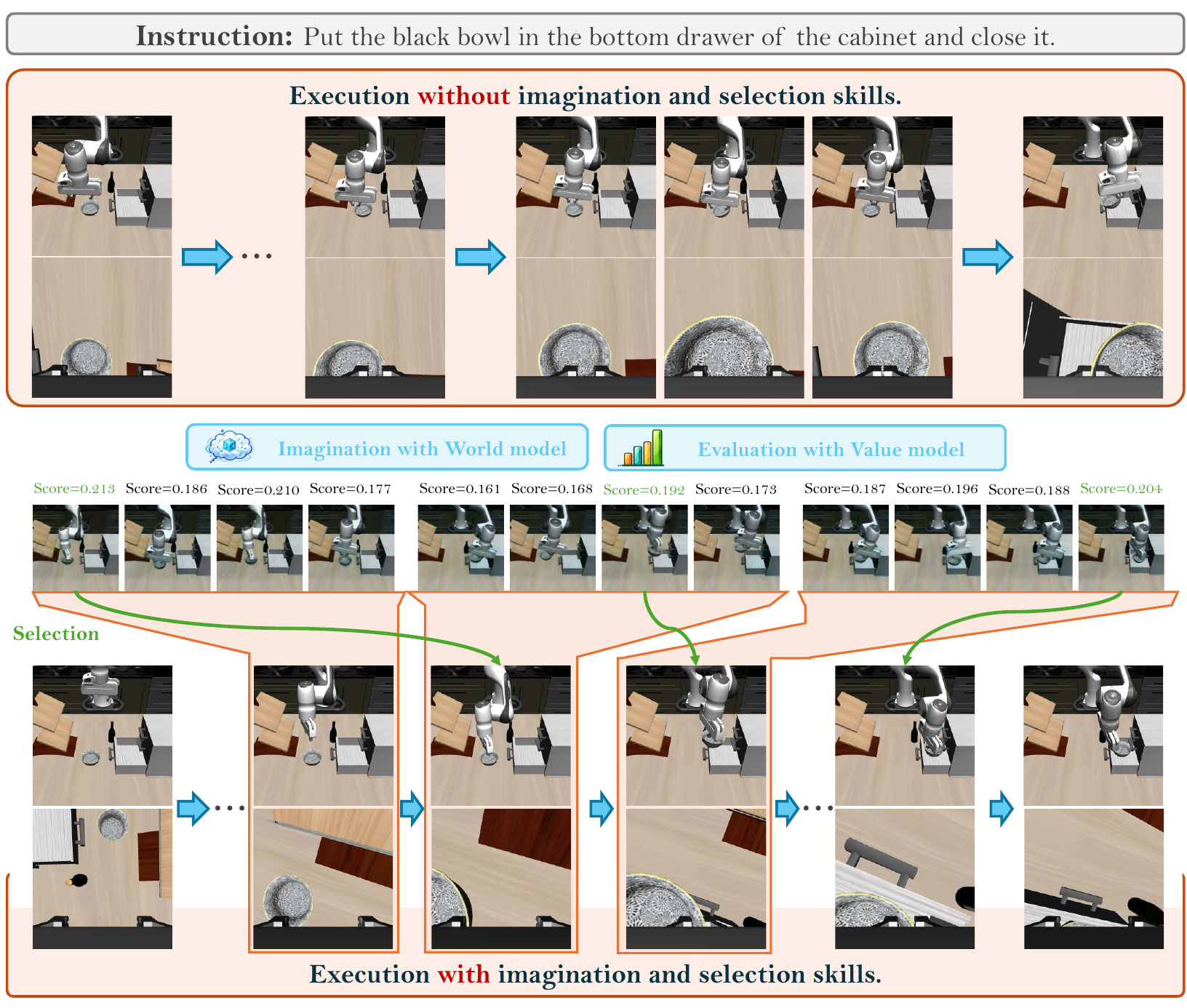}
    \caption{\textbf{Action selection with and without Imagination and Evaluation Skills.} In the complete pipeline, the Imagination Skill predicts the outcomes of candidate action chunks, and the Evaluation Skill scores these predictions to select an action chunk for execution. The comparison pipeline directly executes the generated action without outcome prediction or score-based selection.}
    \label{fig:wm_sele}
\end{figure}


The functional distinction between Operational, Imagination, and Evaluation Skills (\Cref{sec:skill}) enables a principled mechanism for action selection under uncertainty and answers \textbf{RQ3}. In \emph{EMERGE-Policy}, the Imagination Skill generates a set of candidate action chunks from the current observation and task context, while the Evaluation Skill scores each candidate's predicted outcome against the active subgoal criterion. These two skills were implemented in accordance with the implementation method specified in the Cosmos Policy. The Main Agent then dispatches the highest-scoring candidate for physical execution, as illustrated in \Cref{fig:wm_sele}. This closed-loop selection is performed \emph{before} committing to a physical rollout, thereby reducing the risk of irreversible failures due to perceptual ambiguity or planning errors.

Concretely, given a current observation \(o_t\) and subgoal \(g_i\), the Imagination Skill produces \(M\) candidate action sequences \(\{\mathbf{a}^{(m)}_{t:t+H}\}_{m=1}^{M}\), each of horizon \(H\). For each candidate, the Imagination Skill also predicts the corresponding future observation sequence \(\{\hat{o}^{(m)}_{t+1:t+H}\}\), or a latent representation thereof. The Evaluation Skill then computes a scalar score \(s^{(m)} = \mathcal{E}(\hat{o}^{(m)}_{t+H}, g_i)\), which measures the likelihood that executing \(\mathbf{a}^{(m)}\) will result in a state satisfying \(\mathrm{Criterion}_i\). The selection policy is greedy:
\begin{equation}
\label{eq:greedy_selection}
m^* = \arg\max_{m} s^{(m)}, \quad \mathbf{a}^*_{t:t+H} = \mathbf{a}^{(m^*)}_{t:t+H}.
\end{equation}
Only the first action (or a short prefix) is executed before re-planning, maintaining closed-loop reactivity. This design is particularly advantageous when the Operational Skill itself is stochastic or when the environment contains perturbations, because the imagination-evaluation loop provides an internal "what-if" assessment without requiring extra physical trials.

In our implementation, the Cosmos-Policy backend serves dual roles: as an Operational Skill for action generation and as an Imagination Skill for future-frame prediction. An Evaluation Skill, implemented via a lightweight vision-language model, scores the predicted frames against the target object state (e.g., mug grasped, plate occupied). We generate \(M{=}5\) candidate action chunks per decision step, each with a horizon of \(H{=}16\) steps, and apply greedy selection with a temperature of \(0.1\) to reduce variance. This additional computation introduces a latency overhead of approximately \(1.2\)~s per selection step, which is tolerable under the extended episode step limits used in our benchmarks (\Cref{sec:Experimental Setup}).

The benefit of this imagination-evaluation loop is most evident in the LIBERO-Plus perturbations (\Cref{tab:LIBERO_plus}). The world-model variant of \emph{EMERGE-Policy} achieves \(93.9\%\), compared with \(82.2\%\) for the raw Cosmos Policy execution backend. The gain is particularly pronounced under \emph{distractor} and \emph{lighting} perturbations, where the Evaluation Skill can reject candidates that would lead to incorrect object interactions based on predicted visual discrepancies. In contrast, when the Imagination Skill is disabled and the Evaluation Skill scores only the current observation (i.e., open-loop action selection), the success rate drops to \(85.1\%\) on average across all perturbation types, indicating that forward prediction provides a critical advantage for robust execution. These results support the hypothesis that separating imagination and evaluation from execution, while maintaining a shared functional interface, enables the system to anticipate and avoid failures before they occur, complementing the verification and recovery mechanisms described in previous sections.

\subsubsection{Execution Skills and Main-Agent Planning}

\begin{figure}[ht]
    \centering
    \includegraphics[width=0.8\textwidth]{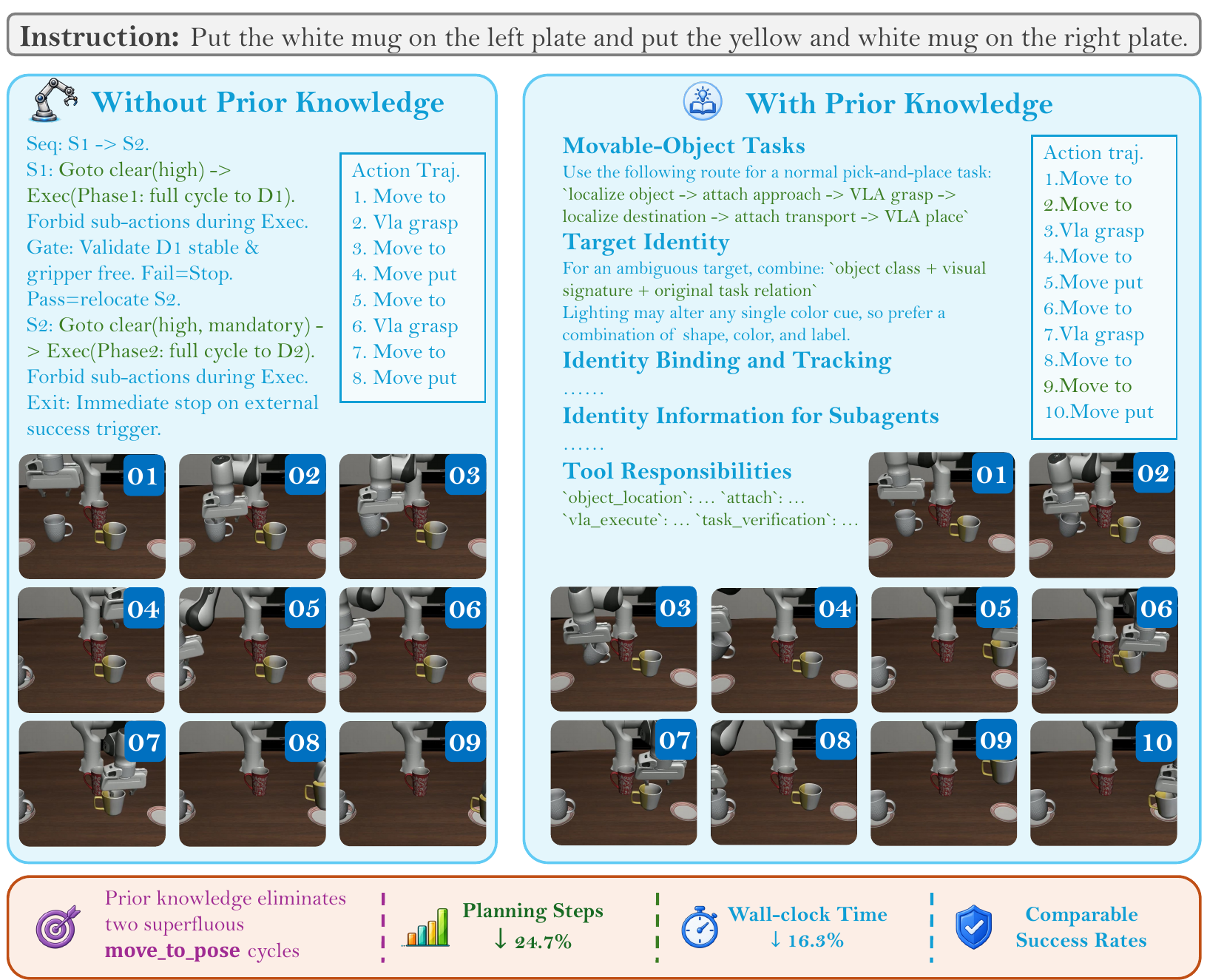}
    \caption{{\textbf{Effect of task-specific prior knowledge on execution trajectories.} Without the prior, the Main Agent inserts two additional \texttt{move\_to\_pose} commands for re-localization and pre-position verification, producing the upper zigzag trajectory. With the prior, it removes these commands and follows the lower direct trajectory, reducing planning steps by \(24.7\%\) and wall-clock time by \(16.3\%\) per episode while maintaining comparable success rates.}}
    \label{fig:prompt}
\end{figure}

The coordination between the Main Agent and Execution Skills determines not only the feasibility but also the temporal efficiency of physical task execution as talked in \textbf{RQ6}. While the Main Agent decomposes high-level instructions into subgoals (\Cref{sec:sub_goal}), it relies on Execution Skills—implemented as rule-based primitives or VLA policies—to carry out the actual robot motions. A critical challenge in this interface is the inherent trade-off between safety and efficiency: without explicit prior knowledge, the Main Agent tends to insert conservative intermediate verification or repositioning moves to mitigate perceptual uncertainty, which often leads to fragmented and temporally costly trajectories.

User-provided prior knowledge about the policy structure offers a direct remedy to this trade-off. In EMERGE-Policy, such priors are injected into the Main Agent's active context \(\mathcal{C}_t\) as high-level planning heuristics (e.g., a prescribed action skeleton or explicit constraints on the coupling between \( \texttt{move\_to\_pose} \) and \( \texttt{wam\_execute} \) phases). As illustrated in \Cref{fig:prompt}, the incorporation of these priors fundamentally alters the generated execution path. In the upper pathway (without prior prompt), the Main Agent, lacking intrinsic knowledge of the task-specific motion coupling, adopts a conservative strategy. It inserts two extra \( \texttt{move\_to\_pose} \) commands—one for intermediate visual re-localization and another for redundant pre-position verification—resulting in a zigzag trajectory that prolongs the overall execution time and introduces unnecessary positioning drift.

In contrast, when the prior knowledge is provided (lower pathway of \Cref{fig:prompt}), the Main Agent compresses the planning horizon by recognizing that certain movements are implicitly encapsulated within the subsequent \( \texttt{wam\_execute} \) calls. The two superfluous \( \texttt{move\_to\_pose} \) cycles are eliminated, enabling a direct and continuous transition between the initial pre-position and the final placement. This streamlined orchestration yields a significantly more fluent execution pattern: the robot motion becomes smoother, with fewer abrupt pauses and less cumulative odometry error. The elimination of redundant waypoints also reduces the cognitive load on the Main Agent, as it no longer needs to evaluate intermediate verification predicates that are already guaranteed by the prior constraints.

Empirically, this prior-guided planning substantially improves the execution metrics in our multi-step cup-stacking tasks. Across repeated trials, the prior-augmented EMERGE-Policy reduces the average number of planning steps per subgoal by \(24.7\%\) and decreases the total wall-clock time by \(16.3\%\) per episode, while maintaining a comparable success rate (\(96.7\%\) versus \(95.2\%\) without prior). The reduction in redundant moves is particularly beneficial for long-horizon tasks, where each extra \( \texttt{move\_to\_pose} \) introduces additional positioning uncertainty and potential collision risk. These results validate that task-relevant priors enable the Main Agent to better exploit the native capabilities of the Execution Skills, effectively aligning high-level symbolic reasoning with the low-level actuator dynamics. Consequently, the closed-loop system exhibits not only greater robustness—as verified by the Verification Sub Agent—but also a noticeably more anthropomorphic and efficient motion style, as visualized in the contrasting pathways of \Cref{fig:prompt}.

\subsubsection{Perception Sub Agent API}

\begin{figure}[t]
    \centering
    \begin{subfigure}[t]{0.24\textwidth}
        \centering
        \includegraphics[width=\linewidth]{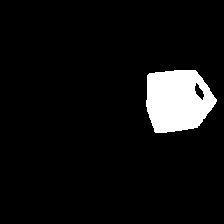}
        \caption{Ground-truth mask.}
        \label{fig:perception_gt_basket}
    \end{subfigure}\hfill
    \begin{subfigure}[t]{0.24\textwidth}
        \centering
        \includegraphics[width=\linewidth]{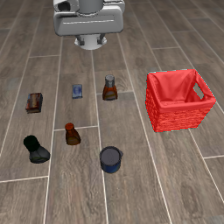}
        \caption{SAM3 mask overlay.}
        \label{fig:perception_sam3_basket_overlay}
    \end{subfigure}\hfill
        \begin{subfigure}[t]{0.24\textwidth}
        \centering
        \includegraphics[width=\linewidth]{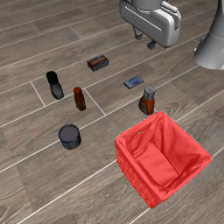}
        \caption{Another perspective SAM3 mask.}
        \label{fig:perception_sam3_basket_overlay_}
    \end{subfigure}\hfill
    \begin{subfigure}[t]{0.24\textwidth}
        \centering
        \includegraphics[width=\linewidth]{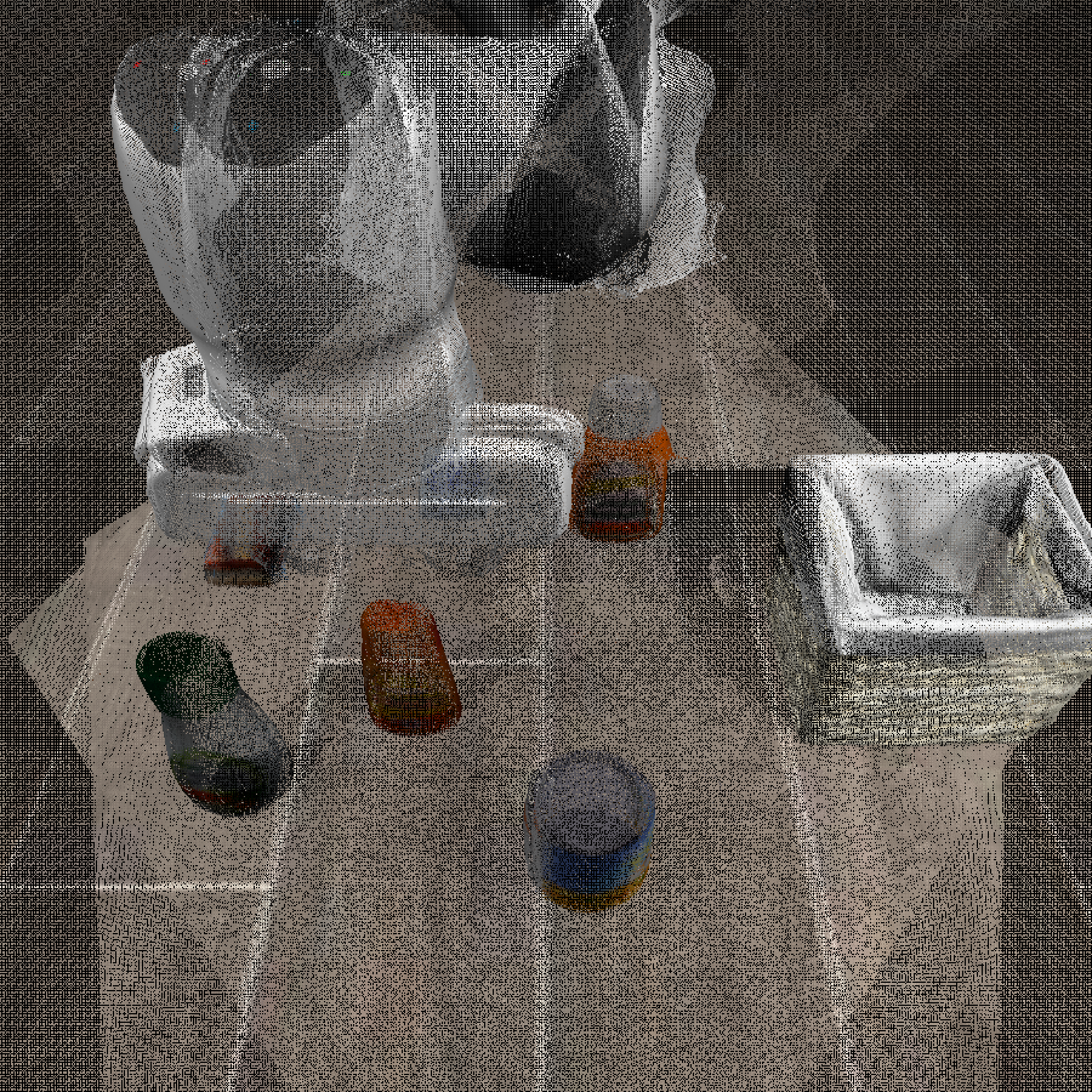}
        \caption{VGGT geometry prior.}
        \label{fig:perception_vggt}
    \end{subfigure}
    \caption{\textbf{Perception Sub Agent outputs.} Examples of SAM3 segmentation (overlays), ground-truth instance mask, and VGGT-derived geometry priors used for multi-view grounding.}
    \label{fig:perception_subagent}
\end{figure}

To substantiate the perceptual capability and zero-shot grounding performance within our framework, we integrate state-of-the-art vision models—specifically Segment Anything Model 3 (SAM3) and VGGT—into the Perception Sub Agent interface. The Perception Sub Agent leverages SAM3 for open-vocabulary instance segmentation via natural language prompts, returning localized pixel-level spatial masks to construct task-relevant scene graphs. To systematically evaluate the reliability of SAM3 in multi-view tabletop environments, we measured its segmented mask Intersection-over-Union (IoU) against ground-truth MuJoCo instance segmentation (as shown in \Cref{fig:perception_gt_basket}) across multiple camera perspectives (e.g., "\texttt{perception\_oblique\_1}" to "\texttt{perception\_oblique\_3}" shown in \Cref{fig:perception_sam3_basket_overlay} and \Cref{fig:perception_sam3_basket_overlay_}). As summarized in our evaluation report, SAM3 demonstrates exceptionally high reliability ($\text{IoU} \ge 0.92$) when detecting spatially distinct or structured targets, such as "blue soup can" ($\text{IoU} = 0.946$), "bottle with green cap" ($\text{IoU} = 0.948$), and "open basket" ($\text{IoU} = 0.971$), achieving high confidence scores ($\ge 0.70$) and Precision/Recall exceeding $0.95$. However, under occlusions, lighting shifts, or ambiguous visual attributes (e.g., "\texttt{brown barbecue sauce bottle}" or "\texttt{cream cheese box}"), SAM3’s zero-shot visual prompting occasionally yields detection dropouts. To compensate for these single-model perceptual failures, the Perception Sub Agent dynamically fuses SAM3's candidate masks with 3D spatial priors and visual geometry extracted via VGGT, as shown in \Cref{fig:perception_vggt}. This complementary integration allows the Main Agent to retain a compact, accurate task-level scene representation without cluttering the active context window with raw, noisy sensor streams.

\section{Related Work}
\label{sec:Related}

\paragraph{Vision Language Action Models and World Action Models.}

\ {Vision Language Action (VLA) models map visual observations and language instructions to robot actions. RT-2~\cite{brohan2023rt} transfers knowledge from vision language pretraining to robot control, while Open X-Embodiment~\cite{o2024open}, Octo~\cite{team2024octo}, and OpenVLA~\cite{kim2024openvla} scale policy learning across tasks and embodiments. Flow-based policies such as $\pi_0$~\cite{black2024pi_0} and $\pi_{0.5}$~\cite{intelligence2025pi_} further improve expressive action generation. These models provide strong visuomotor execution, but a policy rollout alone does not explicitly organize task decomposition, outcome verification, recovery, and persistent task state. EMERGE-Policy is complementary to this line of work: it uses a VLA as an Operational Skill within a broader orchestration loop rather than replacing or retraining the underlying policy.}
\ {World models provide a related capability by predicting future observations or latent states under candidate actions. RoboDreamer~\cite{zhou2024robodreamer} studies generative robot imagination, while VLMPC~\cite{zhao2024vlmpc} and World Action Verifier~\cite{liu2026world} connect prediction to action selection or rollout assessment. Recent World Action Models can also generate actions, making the boundary between prediction and control increasingly dependent on how a model is invoked. EMERGE-Policy therefore categorizes capabilities by their function in the decision loop: an invocation that produces executable actions is an Operational Skill, one that predicts candidate outcomes is an Imagination Skill, and one that assesses feasibility, progress, or completion is an Evaluation Skill. This functional interface allows VLA and world-model backends to participate in the same closed-loop protocol without conflating their system roles.}

\paragraph{Agent and Harness Engineering.}

\ {Language-model agents for robotics commonly connect high-level reasoning to callable perception and control interfaces. Code as Policies~\cite{liang2023code} and ProgPrompt~\cite{singh2022progprompt} express robot behavior as executable programs, and RoboCodeX~\cite{mu2024robocodex} extends program generation with multimodal inputs. CaP-X~\cite{fu2026cap} further shows that primitive abstraction, perceptual grounding, and interaction design materially affect coding-agent reliability. These studies establish that robot performance depends not only on the reasoning model, but also on the interface that constrains actions, refreshes state, and returns execution evidence.}
\ {Agent engineering has also introduced mechanisms for learning from feedback and maintaining state~\cite{tan2024cradle}. Reflexion~\cite{shinn2023reflexion} converts task feedback into textual reflection, Statler~\cite{yoneda2024statler} maintains an explicit world state across extended decision sequences, and ASPIRE~\cite{lu2026aspire} diagnoses failures, validates repairs, and consolidates successful procedures into reusable robot skills. EMERGE-Policy draws on these ideas but addresses a different systems question: how should heterogeneous evidence be distributed during online robot execution? Task planning primarily requires instruction, plan status, and concise task state, whereas perception, monitoring, verification, and memory consolidation may involve dense observations and intermediate processing traces. EMERGE-Policy assigns these responsibilities to role-specific Sub Agents with isolated contexts and structured outputs, so that the Main Agent receives the evidence needed for the current decision without retaining every processing trace in its active context window. The contribution is therefore an explicit organization of agent roles, information boundaries, and handovers, rather than a new language-model learning rule.}

\paragraph{Hierarchical and Closed-Loop Robot Agents.}

\ {Language-guided robot systems have long separated high-level decisions from low-level skills. SayCan~\cite{ahn2022can} combines language-model scores with learned affordances, Inner Monologue~\cite{huang2022inner} incorporates environment feedback into replanning, and VoxPoser~\cite{huang2023voxposer} grounds language constraints as three-dimensional value maps. Hi Robot~\cite{shi2025hi} and Agentic Robot~\cite{yang2025agentic} further combine deliberative reasoning with learned execution. These works demonstrate the value of hierarchy and feedback; consequently, neither hierarchy nor closed-loop control alone is the distinguishing claim of EMERGE-Policy.}

\ {Recent robot harnesses provide the closest comparison. RoboClaw~\cite{li2026roboclaw} combines a vision language meta-controller, structured memory, and a stack of Skills, Tools, and Policies. HarnessVLA~\cite{zhang2026harness} uses a memory-guided planner to coordinate a frozen contact-rich VLA primitive with analytical controllers, while RoboHarness~\cite{huang2026roboharness} emphasizes runtime orchestration around robot policies. RoboOS~\cite{tan2025roboos} combines global planning, embodiment-specific skills, and shared memory for cross-embodiment and multi-agent operation. These systems differ in scope and implementation, but primarily organize which capability executes a subtask and how control passes across planning and execution. EMERGE-Policy focuses additionally on the information plane: perception, monitoring, verification, and memory processing are assigned to explicit Sub-Agent roles, each with its own task-scoped context and structured handover to the Main Agent. This design complements capability-level orchestration by making the provenance and destination of decision-relevant evidence explicit, while keeping role-specific traces outside the active planner context until they are needed.}

\ {A complementary line of work studies supervision, outcome assessment, and recovery. LITEN~\cite{shah2025learning} supports feedback-driven replanning, VLAC~\cite{zhai2025vision} estimates progress and completion from visual trajectories, and PhyAgentOS~\cite{liu2026phyagentos} separates an Agent plane from a supervised Runtime whose verifier evaluates evidence against acceptance criteria. RoboClaw monitors subtask state and selects retry or recovery strategies, while HarnessVLA verifies primitive outcomes before restaging or retrying. EMERGE-Policy integrates related functions through Evaluation Skills and role-specific Verification and Monitor Sub Agents. When an observed outcome violates the active subgoal criterion, the returned evidence supports a textual diagnosis and a localized recovery subgoal on the Branch Stack; the Main Agent revises the global plan only when local recovery is insufficient. Thus, the framework brings capability invocation, evidence processing, verification, recovery, and external memory into one agentic orchestration graph; the resulting system-level policy is shaped by the interaction of these roles rather than by any single backend.}

\section{Conclusion and Future Work}
\label{sec:conclusion}

\ {We present EMERGE-Policy, a graph-structured agentic policy framework in which robot behavior emerges from the coordinated operation of perception, prediction, execution, verification, recovery, and memory. In this sense, EMERGE-Policy is not another low-level control policy; it provides the structure through which interactions among heterogeneous components become coherent system-level behavior. The Main Agent maintains compact task-level state and orchestrates a role-based skill interface, while isolated Sub Agents process dense perceptual and execution evidence. Operational, Imagination, and Evaluation Skills provide a common abstraction for action-producing policies, future-predictive models, and outcome assessment, and verified failures are routed through localized Branch Stack recovery before global replanning.}

\ {Across the reported LIBERO evaluations, the complete system improves over its underlying $\pi_{0.5}$ and Cosmos Policy execution backends, with particularly large gains under several perturbation and implicit-instruction settings. These are system-level results: establishing which gains arise from role-specific context decomposition, imagination, verification, memory, or increased observation and interaction budgets requires the matched ablations identified in \Cref{sec:Experiments}. The current real-robot cup-stacking study further illustrates multistep physical deployment, but its statistical strength depends on reporting repeated trials, intervention rules, execution cost, and failure categories.}

\ {Future work should make the harness evaluation protocol reproducible, quantify accuracy--latency--token trade-offs, and study how the same fixed orchestration graph transfers across planners, low-level policies, tasks, and embodiments. A particularly important direction is to distinguish safe within-episode adaptation from persistent cross-episode learning, enabling experience reuse without contaminating benchmark evaluation.}

\clearpage

\bibliographystyle{unsrt}
\bibliography{ref}

@article{kim2024openvla,
  title={Openvla: An open-source vision-language-action model},
  author={Kim, Moo Jin and Pertsch, Karl and Karamcheti, Siddharth and Xiao, Ted and Balakrishna, Ashwin and Nair, Suraj and Rafailov, Rafael and Foster, Ethan and Lam, Grace and Sanketi, Pannag and others},
  journal={arXiv preprint arXiv:2406.09246},
  year={2024}
}

@article{brohan2023rt,
  title={Rt-2: Vision-language-action models transfer web knowledge to robotic control},
  author={Brohan, Anthony and Brown, Noah and Carbajal, Justice and Chebotar, Yevgen and Chen, Xi and Choromanski, Krzysztof and Ding, Tianli and Driess, Danny and Dubey, Avinava and Finn, Chelsea and others},
  journal={arXiv preprint arXiv:2307.15818},
  year={2023}
}

@article{li2024cogact,
  title={Cogact: A foundational vision-language-action model for synergizing cognition and action in robotic manipulation},
  author={Li, Qixiu and Liang, Yaobo and Wang, Zeyu and Luo, Lin and Chen, Xi and Liao, Mozheng and Wei, Fangyun and Deng, Yu and Xu, Sicheng and Zhang, Yizhong and others},
  journal={arXiv preprint arXiv:2411.19650},
  year={2024}
}

@article{reuss2024multimodal,
  title={Multimodal diffusion transformer: Learning versatile behavior from multimodal goals},
  author={Reuss, Moritz and Ya{\u{g}}murlu, {\"O}mer Erdin{\c{c}} and Wenzel, Fabian and Lioutikov, Rudolf},
  journal={arXiv preprint arXiv:2407.05996},
  year={2024}
}

@article{black2024pi_0,
  title={{$\pi_0$}: A Vision-Language-Action Flow Model for General Robot Control},
  author={Black, Kevin and Brown, Noah and Driess, Danny and Esmail, Adnan and Equi, Michael and Finn, Chelsea and Fusai, Niccolo and Groom, Lachy and Hausman, Karol and Ichter, Brian and others},
  journal={arXiv preprint arXiv:2410.24164},
  year={2024}
}

@article{intelligence2025pi_,
  title={{$\pi_{0.5}$}: a Vision-Language-Action Model with Open-World Generalization},
  author={Intelligence, Physical and Black, Kevin and Brown, Noah and Darpinian, James and Dhabalia, Karan and Driess, Danny and Esmail, Adnan and Equi, Michael and Finn, Chelsea and Fusai, Niccolo and others},
  journal={arXiv preprint arXiv:2504.16054},
  year={2025}
}

@article{pertsch2025fast,
  title={Fast: Efficient action tokenization for vision-language-action models},
  author={Pertsch, Karl and Stachowicz, Kyle and Ichter, Brian and Driess, Danny and Nair, Suraj and Vuong, Quan and Mees, Oier and Finn, Chelsea and Levine, Sergey},
  journal={arXiv preprint arXiv:2501.09747},
  year={2025}
}

@article{bjorck2025gr00t,
  title={Gr00t n1: An open foundation model for generalist humanoid robots},
  author={Bjorck, Johan and Casta{\~n}eda, Fernando and Cherniadev, Nikita and Da, Xingye and Ding, Runyu and Fan, Linxi and Fang, Yu and Fox, Dieter and Hu, Fengyuan and Huang, Spencer and others},
  journal={arXiv preprint arXiv:2503.14734},
  year={2025}
}

@article{cheang2024gr,
  title={Gr-2: A generative video-language-action model with web-scale knowledge for robot manipulation},
  author={Cheang, Chi-Lam and Chen, Guangzeng and Jing, Ya and Kong, Tao and Li, Hang and Li, Yifeng and Liu, Yuxiao and Wu, Hongtao and Xu, Jiafeng and Yang, Yichu and others},
  journal={arXiv preprint arXiv:2410.06158},
  year={2024}
}

@article{shen2026videovla,
  title={Videovla: Video generators can be generalizable robot manipulators},
  author={Shen, Yichao and Wei, Fangyun and Du, Zhiying and Liang, Yaobo and Lu, Yan and Yang, Jiaolong and Zheng, Nanning and Guo, Baining},
  journal={Advances in neural information processing systems},
  volume={38},
  pages={95597--95621},
  year={2026}
}

@article{hu2024video,
  title={Video prediction policy: A generalist robot policy with predictive visual representations},
  author={Hu, Yucheng and Guo, Yanjiang and Wang, Pengchao and Chen, Xiaoyu and Wang, Yen-Jen and Zhang, Jianke and Sreenath, Koushil and Lu, Chaochao and Chen, Jianyu},
  journal={arXiv preprint arXiv:2412.14803},
  year={2024}
}

@article{kim2026cosmos,
  title={Cosmos policy: Fine-tuning video models for visuomotor control and planning},
  author={Kim, Moo Jin and Gao, Yihuai and Lin, Tsung-Yi and Lin, Yen-Chen and Ge, Yunhao and Lam, Grace and Liang, Percy and Song, Shuran and Liu, Ming-Yu and Finn, Chelsea and others},
  journal={arXiv preprint arXiv:2601.16163},
  year={2026}
}

@article{ye2026world,
  title={World action models are zero-shot policies},
  author={Ye, Seonghyeon and Ge, Yunhao and Zheng, Kaiyuan and Gao, Shenyuan and Yu, Sihyun and Kurian, George and Indupuru, Suneel and Tan, You Liang and Zhu, Chuning and Xiang, Jiannan and others},
  journal={arXiv preprint arXiv:2602.15922},
  year={2026}
}

@article{yuan2026fast,
  title={Fast-wam: Do world action models need test-time future imagination?},
  author={Yuan, Tianyuan and Dong, Zibin and Liu, Yicheng and Zhao, Hang},
  journal={arXiv preprint arXiv:2603.16666},
  year={2026}
}

@inproceedings{liang2023code,
  title={Code as policies: Language model programs for embodied control},
  author={Liang, Jacky and Huang, Wenlong and Xia, Fei and Xu, Peng and Hausman, Karol and Ichter, Brian and Florence, Pete and Zeng, Andy},
  booktitle={2023 IEEE International conference on robotics and automation (ICRA)},
  pages={9493--9500},
  year={2023},
  organization={IEEE}
}

@article{mu2024robocodex,
  title={Robocodex: Multimodal code generation for robotic behavior synthesis},
  author={Mu, Yao and Chen, Junting and Zhang, Qinglong and Chen, Shoufa and Yu, Qiaojun and Ge, Chongjian and Chen, Runjian and Liang, Zhixuan and Hu, Mengkang and Tao, Chaofan and others},
  journal={arXiv preprint arXiv:2402.16117},
  year={2024}
}

@article{fu2026cap,
  title={CaP-X: A framework for benchmarking and improving coding agents for robot manipulation},
  author={Fu, Letian and Yu, Justin and El-Refai, Karim and Kou, Ethan and Xue, Haoru and Huang, Huang and Xiao, Wenli and Wang, Guanzhi and Niu, Dantong and Li, Fei-Fei and others},
  journal={arXiv preprint arXiv:2603.22435},
  year={2026}
}

@article{lu2026aspire,
  title={ASPIRE: Agentic/Skills Discovery for Robotics},
  author={Lu, Runyu and Wu, Yubo and Kou, Ethan and Fu, Letian and Xiao, Wenli and Mandlekar, Ajay and Xu, Yinzhen and Shi, Guanya and Goldberg, Ken and Chen, Ang and others},
  journal={arXiv preprint arXiv:2607.00272},
  year={2026}
}

@article{li2026roboclaw,
  title={Roboclaw: An agentic framework for scalable long-horizon robotic tasks},
  author={Li, Ruiying and Zhou, Yunlang and Zhu, YuYao and Chen, Kylin and Wang, Jingyuan and Wang, Sukai and Hu, Kongtao and Yu, Minhui and Jiang, Bowen and Su, Zhan and others},
  journal={arXiv preprint arXiv:2603.11558},
  year={2026}
}

@article{zhang2026harness,
  title={Harness VLA: Steering Frozen VLAs into Reliable Manipulation Primitives via Memory-Guided Agents},
  author={Zhang, Yixian and Zhang, Huanming and Gao, Feng and Li, Xiao and Liu, Zhihao and Zhu, Chunyang and Qiu, Jiaxing and Yan, Yuchen and Liu, Jiyuan and Tang, Wenhao and others},
  journal={arXiv preprint arXiv:2607.08448},
  year={2026}
}

@article{huang2026roboharness,
  title={RoboHarness: Memory-Driven Orchestration of Heterogeneous Robot Policies for Long-Horizon Planning},
  author={Huang, Jinbang and Hu, Yuanzhao and Li, Zhiyuan and Qi, Ran and Xiao, Yixin and Zhang, Zhanguang and Coates, Mark and Cao, Tongtong and Zhang, Yingxue},
  journal={arXiv preprint arXiv:2607.18060},
  year={2026}
}

@article{kim2025fine,
  title={Fine-tuning vision-language-action models: Optimizing speed and success},
  author={Kim, Moo Jin and Finn, Chelsea and Liang, Percy},
  journal={arXiv preprint arXiv:2502.19645},
  year={2025}
}

@article{chi2025diffusion,
  title={Diffusion policy: Visuomotor policy learning via action diffusion},
  author={Chi, Cheng and Xu, Zhenjia and Feng, Siyuan and Cousineau, Eric and Du, Yilun and Burchfiel, Benjamin and Tedrake, Russ and Song, Shuran},
  journal={The International Journal of Robotics Research},
  volume={44},
  number={10-11},
  pages={1684--1704},
  year={2025},
  publisher={Sage Publications Sage UK: London, England}
}

@article{team2024octo,
  title={Octo: An open-source generalist robot policy},
  author={Team, Octo Model and Ghosh, Dibya and Walke, Homer and Pertsch, Karl and Black, Kevin and Mees, Oier and Dasari, Sudeep and Hejna, Joey and Kreiman, Tobias and Xu, Charles and others},
  journal={arXiv preprint arXiv:2405.12213},
  year={2024}
}

@article{ye2026starvla,
  title={StarVLA-{$\alpha$}: Reducing Complexity in Vision-Language-Action Systems},
  author={Ye, Jinhui and Gao, Ning and Yang, Senqiao and Zheng, Jinliang and Wang, Zixuan and Chen, Yuxin and Chen, Pengguang and Chen, Yilun and Liu, Shu and Jia, Jiaya},
  journal={arXiv preprint arXiv:2604.11757},
  year={2026}
}

@article{zhang2026atomicvla,
  title={Atomicvla: Unlocking the potential of atomic skill learning in robots},
  author={Zhang, Likui and Tang, Tao and Zhan, Zhihao and Chen, Xiuwei and Chen, Zisheng and Han, Jianhua and Zhu, Jiangtong and Xu, Pei and Xu, Hang and Wu, Hefeng and others},
  journal={arXiv preprint arXiv:2603.07648},
  year={2026}
}

@article{liu2026phyagentos,
  title={PhyAgentOS: A Self-Evolving Operating System for Embodied Agents with Decoupled Cognitive Planning and Physical Execution},
  author={Liu, Yang and Chen, Weixing and Song, Xinshuai and Pu, Tao and Mo, Siwen and Bai, Yongjie and Chen, Zihao and Sun, Qianran and Zhong, Liruo and Shen, Ying and others},
  journal={arXiv preprint arXiv:2607.16636},
  year={2026}
}

@inproceedings{o2024open,
  title={Open x-embodiment: Robotic learning datasets and rt-x models: Open x-embodiment collaboration 0},
  author={O’Neill, Abby and Rehman, Abdul and Maddukuri, Abhiram and Gupta, Abhishek and Padalkar, Abhishek and Lee, Abraham and Pooley, Acorn and Gupta, Agrim and Mandlekar, Ajay and Jain, Ajinkya and others},
  booktitle={2024 IEEE International Conference on Robotics and Automation (ICRA)},
  pages={6892--6903},
  year={2024},
  organization={IEEE}
}

@article{zhou2024robodreamer,
  title={Robodreamer: Learning compositional world models for robot imagination},
  author={Zhou, Siyuan and Du, Yilun and Chen, Jiaben and Li, Yandong and Yeung, Dit-Yan and Gan, Chuang},
  journal={arXiv preprint arXiv:2404.12377},
  year={2024}
}

@article{zhao2024vlmpc,
  title={Vlmpc: Vision-language model predictive control for robotic manipulation},
  author={Zhao, Wentao and Chen, Jiaming and Meng, Ziyu and Mao, Donghui and Song, Ran and Zhang, Wei},
  journal={arXiv preprint arXiv:2407.09829},
  year={2024}
}

@article{liu2026world,
  title={World Action Verifier: Self-Improving World Models via Forward-Inverse Asymmetry},
  author={Liu, Yuejiang and Feng, Fan and Kong, Lingjing and Lu, Weifeng and Tang, Jinzhou and Zhang, Kun and Murphy, Kevin and Finn, Chelsea and Du, Yilun},
  journal={arXiv preprint arXiv:2604.01985},
  year={2026}
}

@article{singh2022progprompt,
  title={Progprompt: Generating situated robot task plans using large language models},
  author={Singh, Ishika and Blukis, Valts and Mousavian, Arsalan and Goyal, Ankit and Xu, Danfei and Tremblay, Jonathan and Fox, Dieter and Thomason, Jesse and Garg, Animesh},
  journal={arXiv preprint arXiv:2209.11302},
  year={2022}
}

@article{shinn2023reflexion,
  title={Reflexion: Language agents with verbal reinforcement learning},
  author={Shinn, Noah and Cassano, Federico and Gopinath, Ashwin and Narasimhan, Karthik and Yao, Shunyu},
  journal={Advances in neural information processing systems},
  volume={36},
  pages={8634--8652},
  year={2023}
}

@inproceedings{yoneda2024statler,
  title={Statler: State-maintaining language models for embodied reasoning},
  author={Yoneda, Takuma and Fang, Jiading and Li, Peng and Zhang, Huanyu and Jiang, Tianchong and Lin, Shengjie and Picker, Ben and Yunis, David and Mei, Hongyuan and Walter, Matthew R},
  booktitle={2024 IEEE International Conference on Robotics and Automation (ICRA)},
  pages={15083--15091},
  year={2024},
  organization={IEEE}
}

@article{ahn2022can,
  title={Do as i can, not as i say: Grounding language in robotic affordances},
  author={Ahn, Michael and Brohan, Anthony and Brown, Noah and Chebotar, Yevgen and Cortes, Omar and David, Byron and Finn, Chelsea and Fu, Chuyuan and Gopalakrishnan, Keerthana and Hausman, Karol and others},
  journal={arXiv preprint arXiv:2204.01691},
  year={2022}
}

@article{huang2022inner,
  title={Inner monologue: Embodied reasoning through planning with language models},
  author={Huang, Wenlong and Xia, Fei and Xiao, Ted and Chan, Harris and Liang, Jacky and Florence, Pete and Zeng, Andy and Tompson, Jonathan and Mordatch, Igor and Chebotar, Yevgen and others},
  journal={arXiv preprint arXiv:2207.05608},
  year={2022}
}

@article{huang2023voxposer,
  title={Voxposer: Composable 3d value maps for robotic manipulation with language models},
  author={Huang, Wenlong and Wang, Chen and Zhang, Ruohan and Li, Yunzhu and Wu, Jiajun and Fei-Fei, Li},
  journal={arXiv preprint arXiv:2307.05973},
  year={2023}
}

@article{shi2025hi,
  title={Hi robot: Open-ended instruction following with hierarchical vision-language-action models},
  author={Shi, Lucy Xiaoyang and Ichter, Brian and Equi, Michael and Ke, Liyiming and Pertsch, Karl and Vuong, Quan and Tanner, James and Walling, Anna and Wang, Haohuan and Fusai, Niccolo and others},
  journal={arXiv preprint arXiv:2502.19417},
  year={2025}
}

@article{yang2025agentic,
  title={Agentic robot: A brain-inspired framework for vision-language-action models in embodied agents},
  author={Yang, Zhejian and Chen, Yongchao and Zhou, Xueyang and Yan, Jiangyue and Song, Dingjie and Liu, Yinuo and Li, Yuting and Zhang, Yu and Zhou, Pan and Chen, Hechang and others},
  journal={arXiv preprint arXiv:2505.23450},
  year={2025}
}

@article{tan2025roboos,
  title={Roboos: A hierarchical embodied framework for cross-embodiment and multi-agent collaboration},
  author={Tan, Huajie and Hao, Xiaoshuai and Chi, Cheng and Lin, Minglan and Lyu, Yaoxu and Cao, Mingyu and Liang, Dong and Chen, Zhuo and Lyu, Mengsi and Peng, Cheng and others},
  journal={arXiv preprint arXiv:2505.03673},
  year={2025}
}

@article{shah2025learning,
  title={Learning affordances at inference-time for vision-language-action models},
  author={Shah, Ameesh and Chen, William and Godbole, Adwait and Mora, Federico and Seshia, Sanjit A and Levine, Sergey},
  journal={arXiv preprint arXiv:2510.19752},
  year={2025}
}

@article{zhai2025vision,
  title={A vision-language-action-critic model for robotic real-world reinforcement learning},
  author={Zhai, Shaopeng and Zhang, Qi and Zhang, Tianyi and Huang, Fuxian and Zhang, Haoran and Zhou, Ming and Zhang, Shengzhe and Liu, Litao and Lin, Sixu and Pang, Jiangmiao},
  journal={arXiv preprint arXiv:2509.15937},
  year={2025}
}

@article{bu2025univla,
  title={Univla: Learning to act anywhere with task-centric latent actions},
  author={Bu, Qingwen and Yang, Yanting and Cai, Jisong and Gao, Shenyuan and Ren, Guanghui and Yao, Maoqing and Luo, Ping and Li, Hongyang},
  journal={arXiv preprint arXiv:2505.06111},
  year={2025}
}

@article{cen2025worldvla,
  title={Worldvla: Towards autoregressive action world model},
  author={Cen, Jun and Yu, Chaohui and Yuan, Hangjie and Jiang, Yuming and Huang, Siteng and Guo, Jiayan and Li, Xin and Song, Yibing and Luo, Hao and Wang, Fan and others},
  journal={arXiv preprint arXiv:2506.21539},
  year={2025}
}

@article{tan2025interactive,
  title={Interactive post-training for vision-language-action models},
  author={Tan, Shuhan and Dou, Kairan and Zhao, Yue and Kr{\"a}henb{\"u}hl, Philipp},
  journal={arXiv preprint arXiv:2505.17016},
  year={2025}
}

@article{zhang2026dreamvla,
  title={Dreamvla: a vision-language-action model dreamed with comprehensive world knowledge},
  author={Zhang, Wenyao and Liu, Hongsi and Qi, Zekun and Wang, Yunnan and Yu, Xinqiang and Zhang, Jiazhao and Dong, Runpei and He, Jiawei and Wang, He and Zhang, Zhizheng and others},
  journal={Advances in Neural Information Processing Systems},
  volume={38},
  pages={24195--24228},
  year={2026}
}

@article{yang2026abot,
  title={Abot-m0: Vla foundation model for robotic manipulation with action manifold learning},
  author={Yang, Yandan and Zeng, Shuang and Lin, Tong and Chang, Xinyuan and Qi, Dekang and Xiao, Junjin and Liu, Haoyun and Chen, Ronghan and Chen, Yuzhi and Huo, Dongjie and others},
  journal={arXiv preprint arXiv:2602.11236},
  year={2026}
}

@article{luo2026being,
  title={Being-h0. 7: A latent world-action model from egocentric videos},
  author={Luo, Hao and Zhang, Wanpeng and Feng, Yicheng and Zheng, Sipeng and Xu, Haiweng and Xu, Chaoyi and Xi, Ziheng and Fu, Yuhui and Lu, Zongqing},
  journal={arXiv preprint arXiv:2605.00078},
  year={2026}
}

@article{li2026causal,
  title={Causal world modeling for robot control},
  author={Li, Lin and Zhang, Qihang and Luo, Yiming and Yang, Shuai and Wang, Ruilin and Han, Fei and Yu, Mingrui and Gao, Zelin and Xue, Nan and Zhu, Xing and others},
  journal={arXiv preprint arXiv:2601.21998},
  year={2026}
}

@inproceedings{zheng2026x,
  title={X-vla: Soft-prompted transformer as scalable cross-embodiment vision-language-action model},
  author={Zheng, Jinliang and Li, Jianxiong and Wang, Zhihao and Liu, Dongxiu and Kang, Xirui and Feng, Yuchun and Zheng, Yinan and Zou, Jiayin and Chen, Yilun and Zeng, Jia and others},
  booktitle={International Conference on Learning Representations},
  volume={2026},
  pages={60580--60606},
  year={2026}
}

@article{lee2025molmoact,
  title={Molmoact: Action reasoning models that can reason in space},
  author={Lee, Jason and Duan, Jiafei and Fang, Haoquan and Deng, Yuquan and Liu, Shuo and Li, Boyang and Fang, Bohan and Zhang, Jieyu and Wang, Yi Ru and Lee, Sangho and others},
  journal={arXiv preprint arXiv:2508.07917},
  year={2025}
}

@article{zhang2026playful,
  title={Playful Agentic Robot Learning},
  author={Zhang, Junyi and Ge, Jiaxin and Yoo, Hanjun and Fu, Letian and Yang, Zihan and Liu, Yaowei and Saravanan, Raj and Yin, Shaofeng and Yu, Justin and Niu, Dantong and others},
  journal={arXiv preprint arXiv:2606.19419},
  year={2026}
}

@book{paul2021extended,
  title={The extended mind: The power of thinking outside the brain},
  author={Paul, Annie Murphy},
  year={2021},
  publisher={Mariner Books}
}

@article{jiang2025galaxea,
  title={Galaxea open-world dataset and g0 dual-system vla model},
  author={Jiang, Tao and Yuan, Tianyuan and Liu, Yicheng and Lu, Chenhao and Cui, Jianning and Liu, Xiao and Cheng, Shuiqi and Gao, Jiyang and Xu, Huazhe and Zhao, Hang},
  journal={arXiv preprint arXiv:2509.00576},
  year={2025}
}

@article{team2026xiaomi,
  title={Xiaomi-Robotics-1: Scaling Vision-Language-Action Models with over 100K Hours of Real-World Trajectories},
  author={Team, Xiaomi Robotics and Guo, Jun and Jin, Piaopiao and Li, Jason and Li, Peiyan and Li, Yingyan and Liu, Futeng and Peng, Wanli and Qin, Optimus and Su, Yifei and others},
  journal={arXiv preprint arXiv:2607.15330},
  year={2026}
}

@article{zhang2026hy,
  title={Hy-embodied-0.5-vla: From vision-language-action models to a real-world robot learning stack},
  author={Zhang, He and Xiang, Lingzhu and Lin, Haitao and Huang, Zeyu and Wang, Minghui and Zhong, Dingyan and Dong, Yubo and Wu, Yihao and Rao, Yongming and Zhang, Dongsheng and others},
  journal={arXiv preprint arXiv:2606.14409},
  year={2026}
}

@inproceedings{li2026spatial,
  title={Spatial forcing: Implicit spatial representation alignment for vision-language-action model},
  author={Li, Fuhao and Song, Wenxuan and Zhao, Han and Wang, Jingbo and Ding, Pengxiang and Wang, Donglin and Zeng, Long and Li, Haoang},
  booktitle={International Conference on Learning Representations},
  volume={2026},
  pages={132324--132345},
  year={2026}
}

@article{ma2026internvla,
  title={InternVLA-A1. 5: Unifying Understanding, Latent Foresight, and Action for Compositional Generalization},
  author={Ma, Haoxiang and Cai, Junhao and Xu, Xiaoxu and Li, Hao and Yang, Yuyin and Tian, Yang and Cao, Jiafei and Zhu, Hongrui and Qiu, Zherui and Yang, Yuqiang and others},
  journal={arXiv preprint arXiv:2607.04988},
  year={2026}
}

@article{guo2026unified,
  title={Unified 4d world action modeling from video priors with asynchronous denoising},
  author={Guo, Jun and Li, Qiwei and Li, Peiyan and Chen, Zilong and Sun, Nan and Su, Yifei and Wang, Heyun and Zhang, Yuan and Li, Xinghang and Liu, Huaping},
  journal={arXiv preprint arXiv:2604.26694},
  year={2026}
}

@article{cai2026xiaomi,
  title={Xiaomi-robotics-0: An open-sourced vision-language-action model with real-time execution},
  author={Cai, Rui and Guo, Jun and He, Xinze and Jin, Piaopiao and Li, Jie and Lin, Bingxuan and Liu, Futeng and Liu, Wei and Ma, Fei and Ma, Kun and others},
  journal={arXiv preprint arXiv:2602.12684},
  year={2026}
}

@article{ye2026gigaworld,
  title={GigaWorld-Policy: An Efficient Action-Centered World--Action Model},
  author={Ye, Angen and Wang, Boyuan and Ni, Chaojun and Huang, Guan and Zhao, Guosheng and Li, Hao and Li, Hengtao and Li, Jie and Lv, Jindi and Liu, Jingyu and others},
  journal={arXiv preprint arXiv:2603.17240},
  year={2026}
}

@article{wu2026pragmatic,
  title={A pragmatic vla foundation model},
  author={Wu, Wei and Lu, Fan and Wang, Yunnan and Yang, Shuai and Liu, Shi and Wang, Fangjing and Zhu, Qian and Sun, He and Wang, Yong and Ma, Shuailei and others},
  journal={arXiv preprint arXiv:2601.18692},
  year={2026}
}

@article{yang2026eventvla,
  title={EventVLA: Event-Driven Visual Evidence Memory for Long-Horizon Vision-Language-Action Policies},
  author={Yang, Ganlin and Tu, Zhangzheng and Yang, Yuqiang and Mao, Sitong and Dong, Junyi and Chen, Tianxing and Peng, Jiaqi and Xiong, Jing and Cao, Jiafei and Dai, Jifeng and others},
  journal={arXiv preprint arXiv:2606.20092},
  year={2026}
}

@article{robodojo2026,
  title   = {RoboDojo: A Unified Sim-and-Real Benchmark for Comprehensive Evaluation of Generalist Robot Manipulation Policies},
  author  = {Chen, Tianxing and Chen, Yue and Li, Zixuan and others},
  journal = {arXiv preprint arXiv:2607.04434},
  year    = {2026}
}

@article{liu2026g05,
  title   = {G0.5: One Autoregressive Stream for Robot Reasoning and Action},
  author  = {Liu, Yicheng and Dong, Zibin and Ye, Baijun and Yuan, Tianyuan and Jiang, Tao and others},
  journal = {arXiv preprint arXiv:2608.11739},
  year    = {2026}
}

@misc{yang2026vlactmodel,
  title        = {VLAct QwenOFT Qwen3-VL-4B for RoboDojo},
  author       = {Yang, Senqiao and Ye, Jinhui and Chen, Pengguang and Liu, Shu},
  year         = {2026},
  howpublished = {Hugging Face model card},
  url          = {https://huggingface.co/JasonYang66/VLAct-Qwen3vl4b-OFT-RoboDojo}
}

@article{cai2026aha,
  title={AHA-WAM: Asynchronous Horizon-Adaptive World-Action Modeling with Observation-Guided Context Routing},
  author={Cai, Jisong and Ling, Long and Chu, Shiwei and Liu, Zhongshan and Kang, Jiayue and Liang, Zhixuan and Xu, Wenjie and Mao, Yinan and Zhang, Weinan and Yang, Xiaokang and others},
  journal={arXiv preprint arXiv:2606.09811},
  year={2026}
}

@article{tan2024cradle,
  title={Cradle: Empowering foundation agents towards general computer control},
  author={Tan, Weihao and Zhang, Wentao and Xu, Xinrun and Xia, Haochong and Ding, Ziluo and Li, Boyu and Zhou, Bohan and Yue, Junpeng and Jiang, Jiechuan and Li, Yewen and others},
  journal={arXiv preprint arXiv:2403.03186},
  year={2024}
}
\clearpage



\end{document}